\documentclass[11pt]{article}
\PassOptionsToPackage{table}{xcolor}
\usepackage[T1]{fontenc}
\usepackage[utf8]{inputenc} 
\usepackage{textcomp}

\usepackage[letterpaper,margin=1in]{geometry}

\usepackage{times}

\usepackage[protrusion=false,expansion=false]{microtype}

\usepackage{natbib}       
\usepackage{hyperref}
\usepackage{url}

\usepackage{amsmath,amsfonts,bm}

\def\eqref#1{equation~\ref{#1}}

\def\1{\bm{1}}

\DeclareMathAlphabet{\mathsfit}{\encodingdefault}{\sfdefault}{m}{sl}
\SetMathAlphabet{\mathsfit}{bold}{\encodingdefault}{\sfdefault}{bx}{n}

\usepackage{hyperref}
\usepackage{url}
\usepackage{graphicx}
\graphicspath{{figures/}}
\usepackage{subcaption}
\usepackage{wrapfig}
\usepackage{tcolorbox}
\usepackage{longtable}
\usepackage{algorithm}
\usepackage{algpseudocode}

\usepackage{xcolor}
\definecolor{rowgray}{gray}{0.94} 
\usepackage{booktabs}
\usepackage{enumitem}
\usepackage{fvextra} 

\DefineVerbatimEnvironment{cfgblock}{Verbatim}{
  breaklines=true,
  breakanywhere=true,
  xleftmargin=1em
}

\setlist[description]{%
  font=\bfseries,
  style=unboxed,
  nosep,
  topsep=2pt,
  itemsep=0pt,
  parsep=0pt,
  leftmargin=!,       
  labelwidth=3.8cm,   
  labelsep=0.6em
}

\title{Antislop: A Comprehensive Framework for Identifying and Eliminating Repetitive Patterns in Language Models}

\author{%
{\small
\begingroup
\setlength{\tabcolsep}{0pt}
\begin{tabular}{@{}c@{\hspace{1.2em}}c@{\hspace{1.2em}}c@{\hspace{1.2em}}c@{}}
\textbf{Samuel J Paech} &
\textbf{Allen G Roush} &
\textbf{Judah Goldfeder} &
\textbf{Ravid Shwartz\textendash Ziv} \\
{\ttfamily\footnotesize \nolinkurl{spaech@gmail.com}} &
{\ttfamily\footnotesize \nolinkurl{allen.roush@thoughtworks.com}} &
{\ttfamily\footnotesize \nolinkurl{jag2396@columbia.edu}} &
{\ttfamily\footnotesize \nolinkurl{ravidziv@gmail.com}} \\
Independent & Thoughtworks & Columbia University & New York University
\end{tabular}
\endgroup
}%
}

\begin{document}

\maketitle

\begin{abstract}

Widespread LLM adoption has introduced characteristic repetitive phraseology, termed ``slop,'' which degrades output quality and makes AI-generated text immediately recognizable. We present Antislop, a comprehensive framework providing tools to both detect and eliminate these overused patterns. Our approach combines three innovations: (1) \textbf{The Antislop Sampler}, which uses backtracking to suppress unwanted strings at inference time without destroying vocabulary; (2) \textbf{An automated pipeline} that profiles model-specific slop against human baselines and generates training data; (3) \textbf{Final Token Preference Optimization (FTPO)}, a novel fine-tuning method that operates on individual tokens, surgically adjusting logits wherever a banned pattern has appeared in an inference trace. We demonstrate that some slop patterns appear over 1,000$\times$ more frequently in LLM output than human text. The Antislop Sampler successfully suppresses 8,000+ patterns while maintaining quality, whereas token banning becomes unusable at just 2,000. Most importantly, FTPO achieves 90\% slop reduction while maintaining or improving performance in cross-domain evals including GSM8K, MMLU, and creative writing tasks. In contrast, DPO suffers significant degradation in writing quality and lexical diversity despite achieving weaker suppression. We release all code and results under MIT license: \url{https://github.com/sam-paech/auto-antislop}.

\end{abstract}

\section{Introduction}

\begin{wrapfigure}[16]{r}{\dimexpr0.50\textwidth+0.8em\relax}
  \vspace*{-3.5\baselineskip}
  \centering
  \hspace{0.8em}\includegraphics[width=0.50\textwidth]{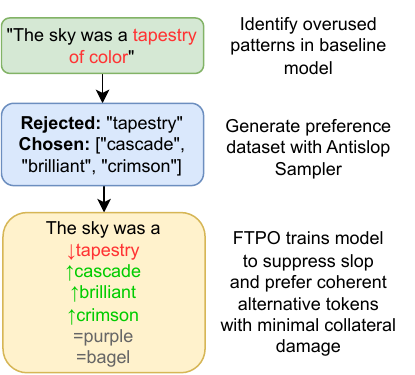}
  \captionsetup{justification=raggedright,singlelinecheck=false,margin={0.8em,0pt}}
  \caption{Pipeline for identifying and suppressing overused writing patterns in a language model.}
  \captionsetup{margin={0pt,0pt}}
  \label{fig:auto-antislop-pipeline}
\end{wrapfigure}

Language models have ushered in an era of slop: Repetitive words and phrases that are instantly recognizable as AI generated text\citep{wu2025survey}. In creative writing, the ubiquitous \textit{Elara} always speaks with "voice barely above a whisper". In functional writing, phrases like \textit{"it's not just X, it's Y"} appear far more frequently than in human text. In our tests, we find that some patterns occur over $1000{\times}$ more frequently in LLM text than in human writing, leading to the perception of repetition and over-use -- i.e. \textit{"slop"}.

Existing approaches to suppress unwanted patterns are brittle or ineffective. Token banning creates collateral damage-- for instance, if we wish to ban "catatonic" and it tokenizes to ["cat", "atonic"], we will have banned all words that tokenize firstly to "cat". Instructing the model to avoid a set of banned vocabulary has limited efficacy and may induce a backfire effect due to the "Pink elephant problem" \citep{castricato2024suppressing}.


We present the Antislop Sampler: it detects unwanted patterns during generation -- words, phrases, and regex patterns -- then backtracks to the pattern's first token, reduces its probability, and resamples. Our sampler can suppress 8,000 patterns with configurable strength (from soft discouragement to hard banning), without degrading output.

To train slop suppression into the model, we present \textbf{Final Token Preference Optimization} (FTPO), a training algorithm designed to surgically suppress slop with minimal collateral damage to the model. Teaching a model to disprefer its \textit{most preferred tokens} requires large logit adjustments, which can damage the model. Our FTPO trainer implements several "soft-touch" mechanisms to minimize deviations from the reference weights. We measure substantial improvements over DPO and token banning on banlist suppression rates, lexical diversity and impact on writing quality.

We release all code and results datasets under MIT license.

\section{Related Work}

Degeneration in text outputs was highlighted by \citet{holtzman-etal-2020-curious}, who showed that maximum-likelihood decoding (e.g. beam search) can lead to bland, looping text. Stochastic decoding strategies like top-$k$, top-$p$ (nucleus sampling), and min-$p$ \citep{nguyen2025turningheatminpsampling} have since been adopted to increase diversity and reduce incoherent outputs. However, these strategies do not address repetitive tendencies in coherent outputs. RLHF has been shown to significantly reduce output diversity compared to a supervised baseline \citep{kirk-etal-2024-rlhf}, and similar effects have been documented for other alignment fine-tuning methods \citep{omahony-etal-2024-mode, murthy-etal-2024-alignment}. Even the use of chat-format templates can suppress creativity, a phenomenon dubbed \textit{diversity collapse} in LLMs \citep{yun-etal-2025-diversity}.

Several recent samplers attempt to improve creativity and diversity while suppressing repetition. \emph{XTC (Exclude Top Choices)} removes the current highest-probability tokens above a threshold \citep{weidmann2024xtc}. This encourages selection of lesser-used continuations, however, it exclusively targets \textit{high probability tokens}, which are not necessarily representative of the model's over-used writing tendencies. \emph{DRY (Don’t Repeat Yourself)} prevents repetition of sequences that have already occurred verbatim in the text multiple times \citep{weidmann2024dry}. This reliably prevents repetitive looping within the current context, however \textit{DRY} is not able to identify repetitive patterns that emerge statistically across a large number of independent generations. ExLlama implements a string banning feature similar to our backtracking mechanism which hard-bans a provided set of strings at inference time \citep{turboderp_exllamav2_0_3_0}.

Beam-search methods exclude forbidden words or phrases by pruning any beam that would produce them. Efficient variants use tries and a fixed beam budget to enforce both positive and negative constraints \citep{hu2019improved}. A recent benchmark compares decoding-time and training-time approaches, and notes that models can still slip around bans with small spelling changes or closely related word forms; they also test simple fixes to reduce this \citep{jon2023negative}.

A similar approach by \citet{zhang-etal-2025-safety} trains a model to deploy a special \texttt{[RESET]} token when unsafe content is detected in the inference trace, triggering backtracking and a retry of the current sentence. Work by \citet{roush-etal-2022-language} further explored lexical filtering at inference time. Their plug-and-play method enforced constraints (such as omitting the letter e in a lipogram) without fine-tuning the model.

 \citet{welleck-etal-2020-neural} introduced an unlikelihood training objective to penalize sequence continuations that exhibit unwanted behaviors (e.g. repetitive loops). This was later generalized by \citet{li-etal-2020-dont} to address dialogue model issues. They added a tailored penalty term to the training loss for disfavored tokens or n-grams.

Our work closely connects to preference-optimization methods like Direct Preference Optimization \citep{rafailov-etal-2023-direct}, which align the model on preference pairs without relying on reward models. However, DPO has known failure modes, including \textit{lowering} the likelihood of preferred responses, inducing diversity collapse and reducing syntactic and n-gram variety in outputs \citep{razin2024unintentional,lanchantin2025divpo,shypula2025diversity}. To counter this, FTPO uses multi-term regularization similar to RLHF's KL penalty \citep{stiennon2020learning}.

\section{Forensic Analysis of Over-represented Patterns}
\subsection{Quantifying Slop}
We identify overused patterns by analyzing statistical overrepresentation of words, bigrams, and trigrams versus human text. For each model, we generate 2,000 outputs using creative writing prompts from Reddit \citep{nitral_reddit_sfw_wp_sharegpt} and compute frequency ratios:
{\setlength{\abovedisplayskip}{6pt plus 2pt minus 2pt}%
 \setlength{\belowdisplayskip}{6pt plus 2pt minus 2pt}%
 \setlength{\abovedisplayshortskip}{4pt}%
 \setlength{\belowdisplayshortskip}{4pt}%
 \[
   \rho(p) = \frac{f_{LLM}(p)}{f_{human}(p)}
 \]
}
where $f_{LLM}(p)$ and $f_{human}(p)$ represent the frequencies of pattern $p$ in LLM and human corpora respectively. Our human baseline combines wordfreq \citep{robyn_speer_2018_1443582} for individual words and a curated corpus of Reddit creative writing and Project Gutenberg texts for n-grams. For n-gram processing, we remove stop-words.

We collate the most overrepresented words and n-grams to produce a "slop fingerprint" of each model's tendencies.

\subsection{Empirical Findings}

Table~\ref{tab:overrep} illustrates the degree of overrepresentation. With \texttt{gemma-3-12b}, certain patterns show extreme usage frequencies compared to human text:

\begin{table}[h]
\centering
\begin{tabular}{lr|lr}
\toprule
\textbf{Word} & \textbf{Ratio} & \textbf{Trigram} & \textbf{Ratio} \\
\midrule
elara & 85,513× & heart hammered ribs & 1,192× \\
unsettlingly & 3,833× & voice trembling slightly & 731× \\
shimmered & 2,882× & said voice devoid & 693× \\
stammered & 2,043× & felt profound sense & 550× \\
\bottomrule
\end{tabular}
\caption{LLM-generated text shows extreme overuse of specific patterns, appearing up to 85,000× more frequently than human writing. The table presents analysis of 2,000 creative writing samples from gemma-3-12b, comparing frequency ratios against human baselines (wordfreq + Reddit/Gutenberg corpora).}
\label{tab:overrep}
\end{table}

The name ``Elara'' appears 85,513 times more frequently in \texttt{gemma-3-12b}'s creative writing outputs than in human text, while the trigram ``heart hammered ribs'' shows 1,192× overrepresentation. Similar overrepresentation ratios are observed in Mistral-small-3.2 and Llama-3-3-70b. Slop fingerprints cluster within model families, but differ between model families (Appendix~\ref{app:slop-clustering}), warranting a model-specific approach to slop identification and suppression.

Our analysis reveals several distinct categories of slop. Models fixate on specific character names (``Elara'', ``Kael''), sensory clichés (``voice barely above a whisper''), intensifiers (``a profound sense'') and a go-to set of overused descriptives (``unsettlingly'', ``shimmered''). We also count sentence-level constructions of the form ``It's not X, it's Y'' to be 6.3× more prevalent than human writing in some models (Figure \ref{fig:notxbuty}).

\section{The Antislop Sampler}

The Antislop Sampler provides inference-time suppression of unwanted patterns. It can suppress individual words (``tapestry''), multi-word phrases (``voice barely above a whisper''), and complex patterns defined by regular expressions (``It's not X, it's Y''). Unlike token banning, which triggers on the first token of a banned sequence and is prone to false positives, our sampler triggers only after the entire sequence appears in the inference trace.
\subsection{Backtracking Mechanism}

\begin{figure}[h]
\begin{center}
\includegraphics[width=0.8\linewidth]{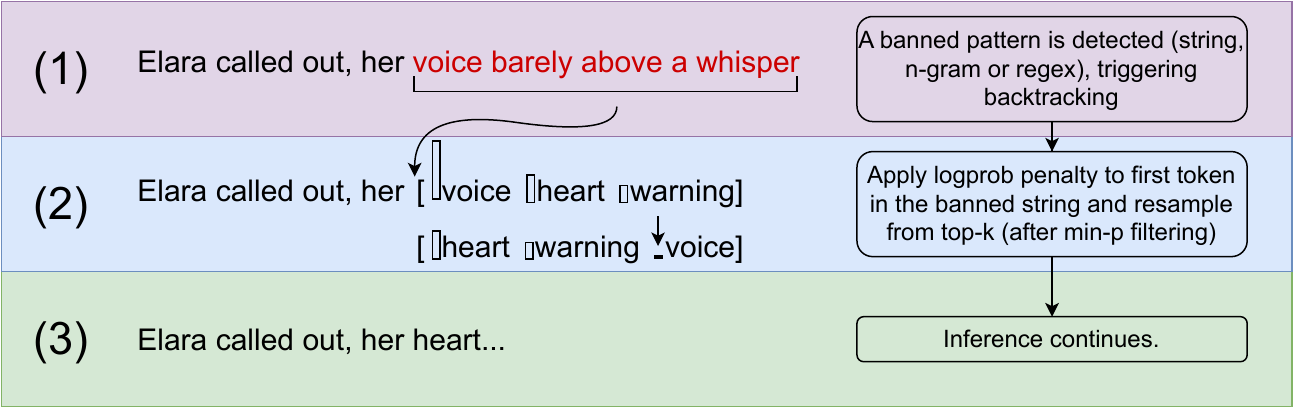}
\end{center}
\caption{The Antislop backtracking mechanism detects unwanted patterns in the inference trace, backtracks to the first token of the banned sequence, lowers its probability, then resamples.}
\end{figure}

During generation, we maintain a trace of tokens and logit distributions, scanning for banned patterns after each token. When detected, we backtrack to the position where the pattern began and lower the initiating token's probability by: $p_{new} = p_{old} \cdot 10^{-10s}$ where $0 \leq s \leq 1.0$ is the configurable \texttt{ban-strength} parameter. We then resample from the adjusted distribution, using min-p filtering to constrain the distribution to coherent candidates meeting a probability threshold. If the same token resamples despite probability reduction, we ignore subsequent violations to prevent infinite loops. This ability to allow banned patterns through if they have sufficiently high probability is a key part of our implementation, which we term "soft-banning".

\begin{algorithm}[H]
\caption{Antislop Backtracking}
\begin{algorithmic}[1]
\ttfamily
\While{generating tokens}
  \State generate token $t$
  \If{banned\_pattern detected}
    \State backtrack to pattern start
    \State reduce probability
    \State resample with min-$p$
  \EndIf
\EndWhile
\end{algorithmic}
\end{algorithm}

\subsection{Soft Banning: Configurable Suppression Strength}
\label{sec:soft_ban}
Imposing a strict ban on words or phrases can cause problems with coherence when there are no viable alternatives. Our soft-banning mechanism provides incremental control through the ban-strength parameter $s$. When $s = 0$, patterns are allowed freely. Values between 0 and 1 provide incremental suppression of the banlist, while $s = 1$ enforces complete blocking.

For example, this approach allows us to generally suppress the word ``tapestry'' while still permitting its use when directly requested in the prompt: ``Write an essay about tapestries''. At \(\texttt{ban-strength < 1.0}\), banned patterns are still allowed through when their probability is high enough compared to the next highest token. See Appendix \ref{app:soft-banning} for a worked example.

\subsection{Implementation and Limitations}

We provide two implementations of the sampler: a single-threaded HuggingFace Transformers version with streaming support, and a multithreaded OpenAI-compatible version for production platforms like vLLM \citep{Kwon2023PagedAttention}.

The sampler suppresses patterns without fine-tuning but reduces throughput. Each backtracking event restarts inference at a prior position, and this may occur hundreds of times per generation with large banlists. In practice, frequent backtracking decreases performance by 69\%-96\%, depending on banlist size (detailed performance analysis in Appendix~\ref{app:performance}). For applications requiring maximum inference speed, this overhead motivates our complementary approach: using the sampler's outputs to train a slop-suppressed model via FTPO.

\section{Final Token Preference Optimization (FTPO)}

We develop Final Token Preference Optimization (FTPO), a training method that permanently suppresses unwanted patterns with minimal degradation to model output. Suppressing slop is nontrivial because it requires large updates to the model's \textbf{most preferred patterns}, reducing their probability until other continuations are preferred. These large shifts can easily damage the model, leading to degradation or model collapse. Our trainer approaches this delicate procedure by incorporating several strategies to constrain logits to the reference, while avoiding collateral damage.

FTPO trains on just a single continuation token at the end of an \textit{incomplete inference trace.} A final-token preference pair consists of three parts:

{
\begin{enumerate}[label=(\arabic*), leftmargin=*, itemsep=2pt, topsep=2pt, parsep=0pt, partopsep=0pt]
  \item The prompt, including the chat template and the model's response up to the point a banned sequence appeared.\\
        \textbf{Prompt:} "\# User: Write me a story. \# Assistant: Once upon a time, Princess"
  \item A single \textit{rejected} continuation token, corresponding to the first token of the banned sequence.\\
        \textbf{Rejected:} \verb!"Elara"!
  \item A set of \textit{chosen} coherent alternative continuation tokens.\\
        \textbf{Chosen:} [\verb!"Madelyne"!, \verb!"Nadia"!, \verb!"Freya"!, \verb!"Isolde"!]
\end{enumerate}
}

\subsection{Limitations of Direct Preference Optimization}

Direct Preference Optimization's (DPO) \citep{rafailov-etal-2023-direct} primary hyperparameter for constraining updates ($\beta$) is a coarse tool, impairing learning at high values and causing model degradation by allowing large logit divergences from reference at small ($\beta$) \citep{wu2024betaDPO}.

Like FTPO, DPO can train on final-token pairs to suppress slop. However, DPO updates only one chosen token per sample, whereas FTPO updates multiple preferred tokens simultaneously.

\subsection{The FTPO Formulation}
\label{sec:ftpo}
FTPO implements several mechanisms to constrain logits to reference, with a two-part regularization allowing larger shifts for \textit{chosen} and \textit{rejected} logits, relative to the remaining vocab. The loss function is formulated as such: At the final position in the inference trace, define token $r$ (rejected) and chosen alternatives $C$. We optimize three loss objectives:

\textbf{Preference loss with margin.} We encourage chosen tokens to exceed the rejected token's logit by margin $m$:
$$\mathcal{L}_{pref} = \frac{\sum_{c \in C} w_c \cdot \text{softplus}(m - \Delta_c)}{\sum_{c \in C} w_c}$$
where $\Delta_c = y[c] - y[r]$ is the logit gap between chosen and rejected, and the weight $w_c = \text{clamp}((m-\Delta_c)/m, 0, 1)$ deactivates the loss when the margin is achieved (Figure \ref{fig:pref-loss-component}).

\textbf{Target regularization.} We tether chosen and rejected ("target") logits to reference values, calculating MSE loss directly on logit deltas (not logprobs). A zero-penalty window $\tau_{target}$ allows these logits initial freedom to move:
$$\mathcal{L}_{target} = \frac{1}{|T|} \sum_{j \in T} \max(|y[j] - y_{ref}[j]| - \tau_{target}, 0)^2$$
where $T = C \cup \{r\}$ contains all target tokens.

\textbf{Non-target regularization.} We strongly anchor the remaining vocabulary to prevent distribution drift:
$$\mathcal{L}_{nontarget} = \frac{1}{|N|} \sum_{j \in N} (y[j] - y_{ref}[j])^2$$
where $N$ represents all non-target tokens.

The total loss, incorporating weighting coefficients $\lambda_{\text{target}}$ and $\lambda_{\text{nontarget}}$:
$$\mathcal{L}_{FTPO} = \mathcal{L}_{pref} + \lambda_{target} \mathcal{L}_{target} + \lambda_{nontarget} \mathcal{L}_{nontarget}$$

\subsection{Key Design Principles}

Three design choices make FTPO effective for targeted suppression of unwanted patterns:

\textbf{Logit-space operation.} With large logit updates to \textit{chosen} and \textit{rejected,} probability mass gets redistributed substantially after softmax, which would impose compensatory pressure on unrelated (non-target) logits if we were to use KL-loss as our regularizer. By using MSE loss on logits instead, we avoid this collateral pressure, localizing updates to just the logits we care about, i.e. the chosen \& rejected.

\textbf{Margin-based deactivation.} The weight $w_c$ automatically reduces to zero when chosen tokens win by margin $m$, preventing overtraining. This self-limiting behavior maintains model stability even with extended training to high preference accuracy.

\textbf{Two-part regularization.} The two-part MSE loss allows target logits to move relatively freely, while constraining the remaining vocabulary to the reference. This allows training to high preference accuracy while avoiding destructive logit divergences.

\subsection{Automated Training Data Generation}

The Antislop Sampler provides an effective mechanism for generating training data for FTPO. At each backtracking event, we capture a preference pair at the exact position where a banned sequence would begin: the \textit{rejected} token that initiated the unwanted pattern versus \textit{chosen} viable alternatives from min-p filtering (Figure \ref{fig:auto-antislop-pipeline}). This enables an end-to-end automated pipeline that identifies overused patterns, generates preference training data, and trains models with FTPO. We release this automated pipeline open-source.

\section{Experimental Evaluation}
\subsection{Experimental Setup}

\textbf{Models.} We evaluate on three model families: Gemma-3-12B, Mistral-Small-3.2, and Llama-3.3-70B, chosen to represent different architectures and scales.

\textbf{Datasets and Benchmarks.} 
For slop analysis and generation, we use creative writing prompts from Reddit \citep{nitral_reddit_sfw_wp_sharegpt}, generating 2,000 samples per model. Human baselines combine wordfreq \citep{robyn_speer_2018_1443582} for word frequencies and curated corpora (Reddit creative writing + Project Gutenberg) for n-gram frequencies.

We evaluate output quality and performance on:

\begin{description}
  \item[MMLU] Multiple-choice STEM and cross-domain knowledge \citep{hendrycks2021mmlu}.
  \item[GSM8K] Generative grade-school math. \citep{cobbe2021gsm8k}.
  \item[Longform Writing] Writing quality is judged by sonnet-4 to a rubric over long (\(\sim\)30k tokens) multi-turn story writing. Particularly sensitive to repetition issues resulting from overtraining. \citep{paech2025longform}.
  
  \item[Writing-Quality Rubric] A GPT-5\textendash judged rubric (Figure~\ref{fig:writing-quality-rubric}) focused on formatting issues, coherence, repetition, and overall quality.
  \item[Diversity] An aggregate, length-controlled lexical diversity metric, normalized to the baseline model at 100. Computed as the mean of MATTR-500 (moving-window TTR, 500-token window), Root-TTR (\(V/\sqrt{N}\)), HD-D (expected unique-word rate from random subsamples), and Distinct-1/2/3 (unique n-grams / total tokens) \citep{Guiraud1960,McCarthyJarvis2010HDD,Li2016Distinct}.
  \item[Banlist Suppression \%] Formulated as the reduction in frequency of banned patterns appearing in outputs, relative to the baseline (0-100\%).
\end{description}

\textbf{Methods Compared.} We evaluate four approaches: (1) token banning with logit bias -100, (2) Antislop Sampler with configurable ban-strength $s$, (3) FTPO fine-tuning, and (4) DPO fine-tuning on identical preference pairs. We test banlist sizes of 2k, 4k, and 8k patterns to assess scalability.

\textbf{Training Details.} Our primary experiments train gemma-3-12b with FTPO and DPO at banlist sizes 2k, 4k and 8k. FTPO uses the hyperparameter configuration specified in Appendix \ref{app:gemma-config}. DPO uses $\beta=0.1$. To minimize perturbation of the original weights, we freeze all layers except the last 5 and lm\_head. We train a high-rank LoRA \citep{hu2021lora} with $r=256$. We find high $r$ allows higher preference accuracy targets to be reached with lower degradation. Both methods train for 1 epoch with early stopping at target preference accuracy of 0.85. For the preference accuracy ablation (\ref{sec:robustness-to-overtraining}), learning rate is scaled such that both methods reached the early stopping targets at approximately the same number of training samples processed.

\subsection{Main Results: Suppression Performance vs. Writing Quality}
\label{sec:suppression-vs-quality}

\begin{figure}[t]
  \centering
  \includegraphics[width=0.7\linewidth]{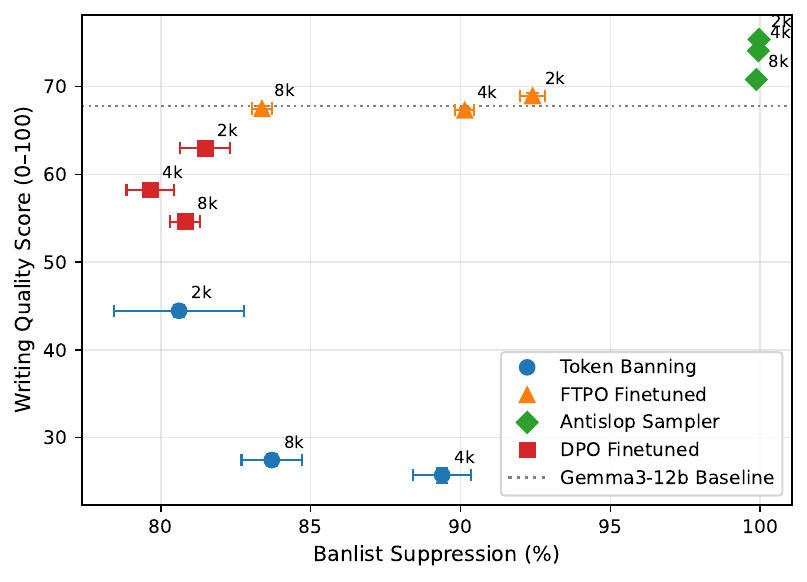}
  \caption{\textbf{FTPO achieves 90\% slop suppression with minimal quality loss, outperforming DPO and token banning.} The figure evaluates four suppression methods on gemma-3-12b across banlist sizes of 2k, 4k, and 8k patterns. FTPO maintains baseline writing quality while suppressing 85-90\% of unwanted patterns. In contrast, DPO degrades quality by 6-15 points despite achieving only 80-82\% suppression. Token banning shows catastrophic quality collapse. Error bars show 95\% confidence intervals (CI\textsubscript{95}). $n=1{,}000$ outputs per condition.}
  \label{fig:quality_vs_banlist_reddit}
\end{figure}

Figure~\ref{fig:quality_vs_banlist_reddit} visualizes the performance in banlist suppression for each method, plotted against output degradation as measured by our writing rubric. The Antislop Sampler achieves perfect suppression (100\%) while actually improving writing quality above baseline. FTPO maintains quality within 1\% of the baseline performance of gemma-3-12b, while achieving 83-92\% suppression rates. 

In contrast, DPO and token banning show marked quality degradation. DPO drops 6-15 points in writing quality despite achieving only 80-82\% suppression. Token banning collapses even more severely, with quality falling to 28 (out of 100) at 8k patterns. In practice, this degradation manifests as severe repetition, spelling and grammar artifacting, and incoherence. These performance disparities demonstrate a clear advantage of Antislop and FTPO over prior methods.

The writing dataset used for evaluation consists of 1,000 prompts from the same Reddit writing dataset \citep{nitral_reddit_sfw_wp_sharegpt} we used in training. For evaluation, we select a subset that excludes these training prompts. We also demonstrate generalization on an out-of-distribution writing dataset, EQ-Bench Creative Writing, with comparable results (Figure \ref{fig:quality_vs_banlist_eqbench}).

\begin{tcolorbox}[colback=gray!5,colframe=black,left=1mm,right=1mm,top=1mm,bottom=1mm]
\textbf{Key Result:} FTPO achieves 90\% suppression with $<1\%$ quality loss, while DPO achieves 80\% suppression with 15\% quality degradation.
\end{tcolorbox}

\subsection{FTPO vs DPO: Detailed Comparison}
FTPO maintains strong suppression across models with minimal degradation (Table~\ref{tab:antislop-ftpo-eval-results}). FTPO suppresses 90+\% of slop for banlist sizes $<= 4,000$ items, with negligible impact on writing quality metrics, lexical diversity and math/STEM benchmarks.

\textbf{Suppression effectiveness.} FTPO achieves 8.5\% stronger suppression than DPO at equivalent training settings. 

\textbf{Capability preservation.} FTPO maintains math reasoning on GSM8k and world-knowledge capabilities on MMLU within 1-3\% of baseline. DPO degrades both metrics by 2-5\%.

\textbf{Long-form generation.} The difference is most dramatic in the longform creative writing test, since repetition and other degradation modes are exacerbated in extended multi-turn generation. Our FTPO-trained models cluster around the baseline gemma3 score for 2k, 4k and 8k banlist sizes; while DPO experiences a large degradation in quality.

\textbf{Lexical diversity.} FTPO maintains or enhances diversity (95-102\% of baseline), while DPO causes progressive collapse (74-92\%). This confirms our hypothesis: DPO has collateral effects on probability distributions, while FTPO's precise adjustments preserve vocabulary diversity.

This pattern generalizes across all evaluated models (12B-70B parameters) and architectures. We note a caveat: Llama-3.3-70B proved more sensitive to preference training, being prone to repetition and degradation artifacts. To mitigate, we restrict LoRA training to $lm\_head$ for this model, resulting in a weaker suppression rate of 66\%.

\setlength\LTleft{0pt}
\setlength\LTright{0pt}

{%
\setlength{\tabcolsep}{4pt} 
\begin{longtable}{lrrrrrr}
\caption{FTPO \& DPO evaluation results for models fine-tuned to suppress a range of banlist sizes from of 1k to 8k patterns. Shown are results on MMLU, GSM8k, Longform Writing, Writing Quality per our rubric, Lexical Diversity (normalized to baseline), and banned pattern suppression rate relative to baseline.\label{tab:antislop-ftpo-eval-results}}\\
\toprule
experiment & mmlu & gsm8k & longform & writing qual & diversity & ban \% \\
\midrule
\endfirsthead

\toprule
experiment & mmlu & gsm8k & longform & writing qual & diversity & ban \% \\
\midrule
\endhead

\midrule
\multicolumn{7}{r}{Continued on next page} \\
\midrule
\endfoot

\bottomrule
\endlastfoot

\rowcolor{rowgray} gemma-3-12b baseline   & 0.590 & 0.888 & 51.3 & 67.80 & 100.00 & 0.00 \\
\rowcolor{rowgray} gemma-3-12b FTPO 2k (Ours)    & 0.559 & 0.876 & 47.5 & \textbf{68.93} & \textbf{101.05} & \textbf{92.39} \\
\rowcolor{rowgray} gemma-3-12b FTPO 4k (Ours)    & 0.565 & 0.880 & 49.4 & 67.31 & 97.68  & 90.15 \\
\rowcolor{rowgray} gemma-3-12b FTPO 8k (Ours)    & \textbf{0.592} & \textbf{0.889} & \textbf{52.3} & 67.49 & 95.09  & 83.40 \\
\rowcolor{rowgray} gemma-3-12b DPO 2k     & 0.541 & 0.847 & 36.6 & 62.98 & 91.03  & 82.00 \\
\rowcolor{rowgray} gemma-3-12b DPO 4k     & 0.549 & 0.861 & 34.8 & 58.24 & 81.92  & 80.64 \\
\rowcolor{rowgray} gemma-3-12b DPO 8k     & 0.571 & 0.864 & 26.9 & 54.61 & 73.92  & 81.44 \\

Mistral-Small baseline & 0.812 & 0.900 & 56.03  & 72.93 & 100.00 & 0.00 \\
Mistral-Small FTPO 1k (Ours)  & 0.811 & 0.895 & 58.38  & 74.60 & 102.10 & 89.46 \\

Llama-3.3-70B baseline     & 0.801   & 0.929 & 36.77  & 64.34 & 100.00 & 0.00 \\
Llama-3.3-70B FTPO 1k (Ours)     & 0.799   & 0.923   & 35.57  & 63.16 & 99.66 & 66.41 \\
\end{longtable}\vspace{-4pt}
}%

\subsection{Robustness to Overtraining}
\label{sec:robustness-to-overtraining}

Compared with DPO, FTPO can train to a higher preference accuracy target on final-token preference pairs before degradation or model collapse occurs. FTPO is designed to precisely alter only the logits needed, switching off the training signal when \textit{chosen} logits are winning by a given margin over \textit{rejected}. DPO lacks these "soft-touch" features, resulting in chosen/rejected logits continuing to diverge as training progresses.

\begin{figure}[h]
\centering
\begin{subfigure}[b]{0.35\linewidth}
  \vspace*{0.0\linewidth}
  \includegraphics[width=\linewidth]{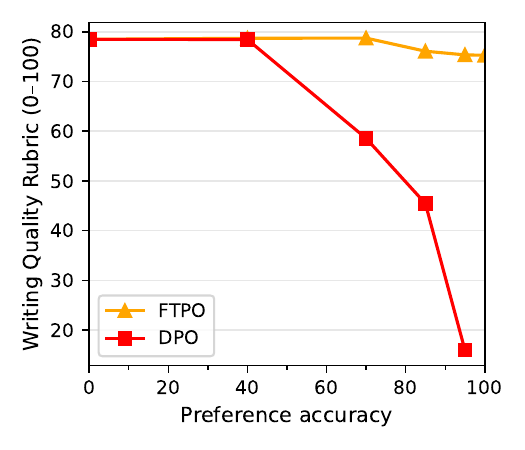}
  \vspace*{-0.085\linewidth}
  \captionsetup{width=.9\linewidth,justification=centering}
  \caption{FTPO maintains writing quality as training progresses to higher pref accuracies, while DPO degrades sharply after the 40\% accuracy mark. This experiment trains gemma-3-12b on a banlist of 1,000 items.}
  \label{fig:overtraining-degradation}
\end{subfigure}\hfill
\begin{subfigure}[b]{0.65\linewidth}
  \includegraphics[width=\linewidth]{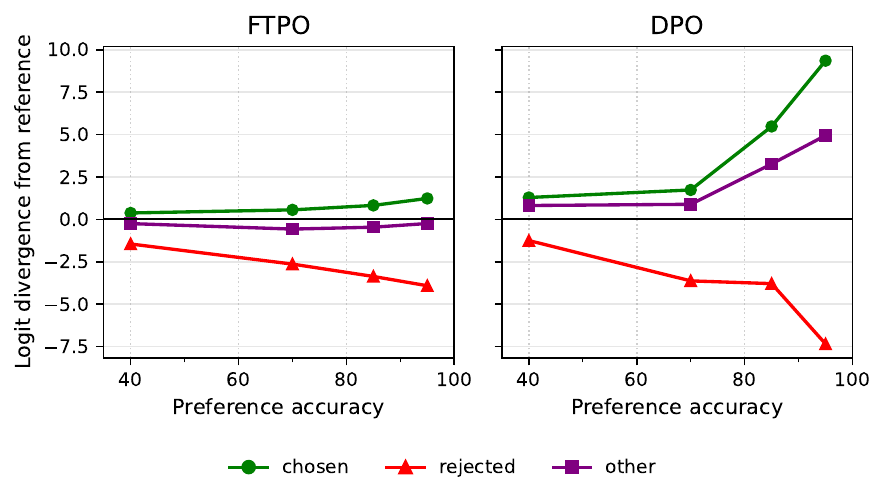}
  \captionsetup{width=.9\linewidth,justification=centering}
  \caption{With FTPO, logits stay close to reference due to (1) the MSE loss terms and (2) the early switch-off feature which  nulls the training signal for chosen tokens that are already winning vs rejected. With DPO, logits diverge unconstrained as training continues. We posit this to be the main cause of FTPO's minimal degradation vs DPO.}
  \label{fig:logit-divergence}
\end{subfigure}
\caption{(a) Impact on writing quality from training to high preference accuracy targets; (b) Logit divergence from reference as training progresses.}
\label{fig:side-by-side}
\end{figure}

When training gemma-3-12b to increasing preference accuracy targets, we find FTPO can train to nearly 100\% preference accuracy with minimal degradation, while DPO only manages 40\%, after which substantial degradation occurs (Figure \ref{fig:overtraining-degradation}). Increasing DPO's $\beta$ hyperparameter to 1.0 mitigates this degradation, but impairs learnability, reducing ban suppression by $15.9\%$ (Figure \ref{fig:dpo-beta-ablation}). We posit that FTPO's mechanisms for constraining logits to the reference while allowing freedom of movement of target logits are the primary reasons it outperforms DPO on this task.



\subsection{Regex Bans}

We perform an experiment to demonstrate suppression of variable sentence-level patterns with regex bans. The Antislop Sampler is able to suppress 100\% of "It's not X, it's Y" patterns (Appendix \ref{app:regex-bans}).

\subsection{FTPO Hyperparameter Ablations}

The FTPO trainer exposes hyperparameters to tune the strength of the MSE loss tether to the reference, and also the margin specifying where gradients turn off for winning \textit{chosen} logits. We train gemma-3-12b on hyperparameter ranges outside the defaults, observing poor preference accuracy and degradation at these sub-optimal values, and thus demonstrating the efficacy of these FTPO safeguards (Appendix \ref{app:hyperparameter-ablations}).

\section{Discussion}

Antislop Sampler achieves 100\% suppression of over-used patterns without quality loss. FTPO outperforms DPO on our measured metrics, even for 30,000-token generations.

Our methods have several limitations: Antislop Sampler reduces throughput by 69-96\% (banlist sizes 1k-8k) due to backtracking frequency. In performance-sensitive deployments, this is a clear incentive to prefer a solution that trains suppression into the weights. 

Anticipating these downstream needs, we develop a pipeline that automatically profiles a model's overused writing patterns, generates a training set, and trains the model to suppress these patterns. Our FTPO trainer is designed to make targeted adjustments to the model's over-used writing tendencies with minimal changes to its distribution otherwise. FTPO's minimal degradation stems from its multi-part regularization and gradient nullification when chosen tokens exceed the margin.

We encourage future work to explore Antislop's performance in domains other than creative writing, human-rater replication of quality metrics, AI generated text detection, and suppression of toxic text.

\section{Conclusion}

We introduced a framework for eliminating overused stylistic patterns ("slop") in LLM outputs while preserving capabilities on our evaluated benchmarks. The \emph{Antislop sampler} performs sequence-level enforcement with a backtracking resample that preserves coherence, supports hard and soft bans, and can suppress  string and regex patterns. Our automated pipeline extracts model-specific slop fingerprints by comparing the model's overused writing patterns against human baselines, then synthesizes a preference dataset without human intervention. \emph{Final Token Preference Optimization (FTPO)} trains the model on these pairs, making suppression permanent. Across our tests, FTPO and the sampler achieved higher suppression than DPO and logit-based token banning, with negligible measurable quality loss on our rubric. We release code and datasets under the MIT license.

\textbf{AI Usage Disclosure:} Language models were used to assist with early drafting of sections of this paper. All results were human designed and performed, and the citations were human-sourced and validated.

\section*{Reproducibility Statement}
We provide all materials to reproduce our results. Algorithms are specified in Sections~\ref{sec:soft_ban}–\ref{sec:ftpo} including loss definitions and hyperparameters. The general configuration template for FTPO/DPO training configuration, LoRA settings, early-stopping criteria, and decoding parameters are given in App.~\ref{app:gemma-config}. In addition, the data pipeline, prompts, judge rubric, and scoring template are included (Fig.~\ref{fig:writing-quality-rubric}). For inference with Antislop, we describe the implementation and throughput (App.~\ref{app:performance}), and include our antislop-vllm implementation in supplementary materials. The supplemental materials contain necessary code and example configuration files to run Antislop Sampler and the automated training pipeline with FTPO or DPO.

\section*{Ethics Statement}
We adhere to the ICLR Code of Ethics (\url{https://iclr.cc/public/CodeOfEthics}). Our study operates on publicly available datasets and benchmarks: Reddit SFW Writing Prompts via Nitral-AI \citep{nitral_reddit_sfw_wp_sharegpt}, EQ-Bench creative prompts \citep{paech2023eqbench}, Project Gutenberg texts \citep{project_gutenberg}, and wordfreq statistics \citep{robyn_speer_2018_1443582}. We processed only public text and did not collect or annotate human subjects. No personally identifying information was collected, and no IRB was required.

Potential harms include: (i) unintended suppression of legitimate dialects, or minority styles; (ii) attempts to evade AI-text detection. Mitigations: our code produces human-readable banlists which may be vetted by hand before deployment; we document and expose the \emph{ban-strength} control (Sec.~\ref{sec:soft_ban}) and provide soft-ban defaults rather than hard blocking; we implement a whitelist to prevent terms from being automatically banned; we recommend human review of any production banlist. Our methods do not target model safety filters and are not intended to bypass them.

We transparently report throughput impacts (App.~\ref{app:performance}) to support energy-cost accounting. The authors declare no conflicts of interest, no external sponsorship that biases results, and disclose LLM assistance for drafting as stated in the paper’s AI Usage Disclosure.

\section*{Acknowledgements}
We thank \textit{Thoughtworks} for generously providing compute for several of our experiments.

\bibliographystyle{plainnat}
\bibliography{iclr2026_conference}

\clearpage
\section*{Appendices}
\appendix

\section{Soft Banning}
\label{app:soft-banning}

In real-world use cases, it is often not preferable to ban a word or phrase outright. In these cases, a scalable "soft ban" is preferred, where there is a general suppression effect, but the suppressed vocab may still be used if there are no good alternatives.

An example of how soft-banning works when there are no good alternate candidates:
\begin{enumerate}[label=Step \arabic*., noitemsep, topsep=0pt]
  \item We have the word ``tapestry'' in our banlist, and have set ban-strength = 0.2 and min-p = 0.1.
  \item The user requests an essay on tapestry weaving.
  \item The model begins inference with, ``The art of Tapestry-'', triggering backtracking. In this example we will say ``Tapestry'' was the top token at this position with 0.99 prob, with the next highest token ``Mural'' at 0.0005.
  \item The ``Tapestry'' token is reduced to \(\mathrm{prob}_{\text{new}} = 0.99 \times 10^{-10\cdot 0.2} = 0.0099\).
  \item After probability rescaling, min-p still excludes ``Mural'' from consideration, since \(\tfrac{0.0005}{0.0099} \approx 0.05 < 0.1\) (the min-p threshold), resulting in ``Tapestry'' remaining the only candidate for sampling.
  \item ``Tapestry'' is selected as the next token despite being on the banlist. This specific violation at this position is marked to be ignored by Antislop in future checks, to avoid a backtracking loop.
\end{enumerate}

A ban-strength value of 1.0 is effectively a hard ban, enforcing 100\% suppression of the banlist.

To determine whether each method can still use the suppressed patterns when contextually necessary, we construct an adversarial prompt:

\begin{list}{}{\leftmargin=1em \rightmargin=1em \topsep=3pt \partopsep=0pt \parsep=0pt \itemsep=0pt}
\item\relax
Write a short story (500 words) incorporating the target phrase exactly 3 times in the story.\\
The target phrase is: “\{phrase\}”.
\end{list}

\begin{figure}[h]
\begin{center}
\includegraphics[width=0.9\linewidth]{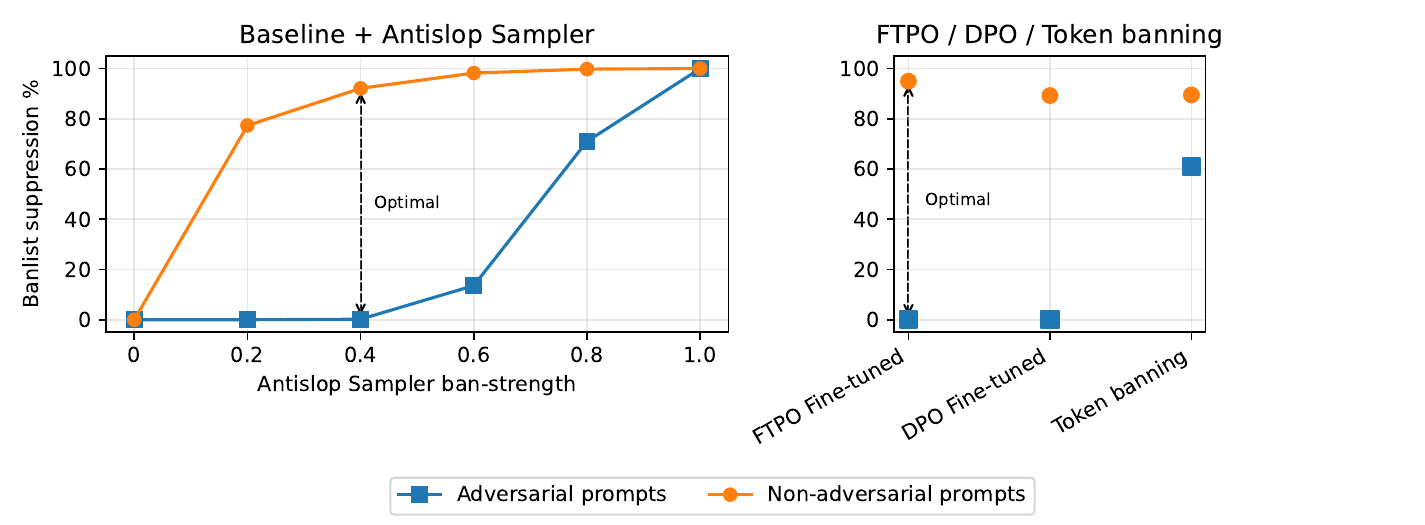}
\end{center}
\caption{Our methods can suppress 90+ percent of banlist occurrences while allowing the banlist through when contextually necessary. Antislop Sampler, FTPO, DPO and token banning are compared on banlist suppression efficacy under normal writing conditions (non-adversarial prompts) and when the model is explicitly instructed to use the banned vocab (adversarial prompts). We indicate optimal behavior for most real-world use cases to be \textbf{maximal suppression in normal writing conditions}, and \textbf{minimal (preferably zero) suppression in adversarial conditions} -- i.e. when the model has no coherent alternatives.}
\label{fig:adversarial-suppression}
\end{figure}

Figure \ref{fig:adversarial-suppression} validates the soft-banning mechanism (Section \ref{sec:soft_ban}), where ban-strength $s$ controls suppression intensity. The Antislop Sampler with $s=0.4$ achieves optimal balance, suppressing patterns in 90\% of normal generation (non-adversarial) while fully permitting them when explicitly requested.


\clearpage
\section{Inference Performance (tok/s)}
\label{app:performance}

We release two implementations of the Antislop sampler: A single-threaded version using Huggingface Transformers, and a higher-throughput version that works with any OpenAI-compatible v1/completion endpoint that supports top\_logprobs. The sampler incurs significant throughput penalty, especially with larger banlist sizes, due to the backtracking events. There is additional performance lost with the API implementation, since it generates in chunks, with banned pattern detection only occurring after a chunk is generated. This could be optimized further by, for example, integrating the sampler into vLLM directly rather than generating chunkwise via the API.

The maximum token rate of our OpenAI API implementation is discovered with binary search on the number of concurrent threads when generating with vLLM. Figures cited are using a single Nvidia H100 gpu.

We measure a 69\% reduction in throughput at a banlist size of 1,000, up to 96\% reduction at banlist size 8,000. However, these should be considered worst-case values. A banlist of this size would be overkill for most real-world usage; we include it here as a stress-test.

\begin{figure}[h]
\begin{center}
\includegraphics[width=0.7\linewidth]{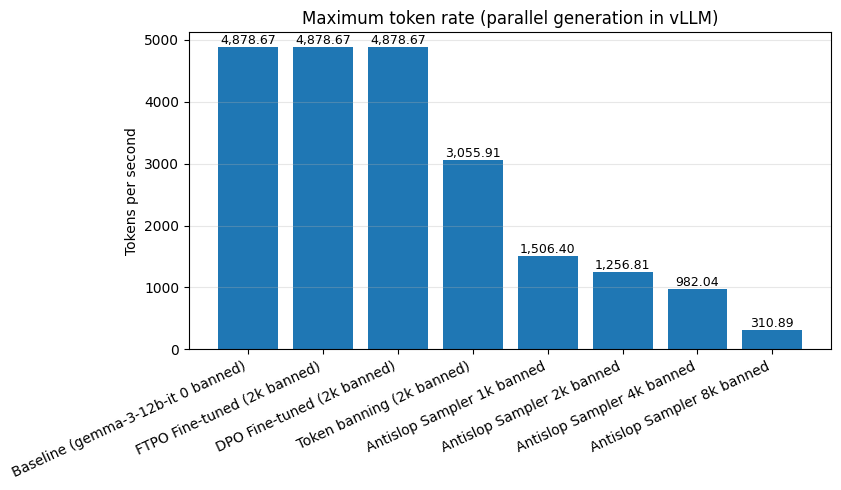}
\end{center}
\caption{Rate of inference is measured for each method when generating with optimal parallelism with vLLM.}
\label{fig:diversity-chart}
\end{figure}

\clearpage
\section{Long-range constraint enforcement via regex bans}
\label{app:regex-bans}

Some models exhibit stylistic slop such as the ``not $x$, but $y$'' family of constructions, which standard quality metrics rarely penalize and which are difficult to unlearn post hoc. We prevent these forms at inference by compiling a small set of regular expressions into one alternation and scanning the full generated text each validation pass. On a match we locate the earliest offending span, map its first character to the corresponding generated-token index, and trigger backtracking at that position. Backtracking resamples from the cached top-logprob lists with the same decoding hyperparameters (temperature, top-$p$, top-$k$, min\_$p$), yielding a coherent alternative continuation without another API call.

\begin{figure}[h]
\begin{center}
\includegraphics[width=0.95\linewidth]{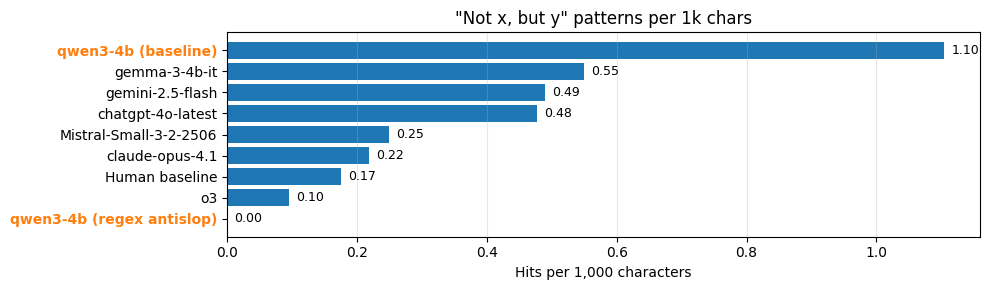}
\end{center}
\caption{Occurrences per 1k characters of the ``not $x$, but $y$'' family across several models. The Antislop variant of \texttt{qwen3-4b} enforces regex bans with backtracking and yields 0.00 hits.}
\label{fig:notxbuty}
\end{figure}

Figure~\ref{fig:notxbuty} shows an example where the baseline \texttt{qwen3-4b} overuses the pattern, while Antislop with regex bans reduces its rate to zero.

\clearpage
\section{Hyperparameter Ablations}
\label{app:hyperparameter-ablations}

The FTPO trainer exposes some tunable hyperparameters:

\textbf{clip\_epsilon\_logits}: Clips the preference-loss component of the training signal for \textit{chosen logits} that are already beating the rejected logit by this margin.

\textbf{lambda\_mse\_target}: The strength of the tethering to reference logits, specifically applied to the target (chosen \& rejected) logits. Higher values prevent the target logits straying too far from reference, but also make it harder for the trainer to achieve high preference accuracy. Lower values allow the model to learn more easily, but may lead to degradation or model collapse.

In this ablation, we train gemma-3-12b with FTPO on 10k samples with early stopping at 95\% preference accuracy. We vary clip\_epsilon\_logits from 2 (default) to 16 while keeping other parameters at defaults, to demonstrate the protective effect of this feature of the trainer. We also ablate the lambda\_mse\_target parameter, setting it at 0, 0.05 (default) and 0.4 while keeping other parameters at defaults. We measure the impact on writing quality, average divergence of logits from reference, and the percent of training examples processed before the 95\% preference accuracy early stopping condition is triggered.

\setlength\LTleft{0pt}
\setlength\LTright{0pt}
\begin{longtable}{lrrrrrrr}
\caption{FTPO ablation results for clip\_epsilon\_logits and lambda\_mse\_target.\label{tab:ftpo-ablation-results}}\\
\toprule
experiment & writing qual & ban \% & early stop & $\Delta$chosen & $\Delta$rejected & $\Delta$other \\
\midrule
\endfirsthead

\toprule
experiment & writing qual & ban \% & early stop & $\Delta$chosen & $\Delta$rejected & $\Delta$other \\
\midrule
\endhead

\midrule
\multicolumn{7}{r}{Continued on next page} \\
\midrule
\endfoot

\bottomrule
\endlastfoot

gemma-3-12b baseline & 67.80 & 0.00 & N/A & N/A & N/A & N/A \\
default params & 67.89 & 84.51 & 66.00 & 1.23 & -3.93 & -0.26 \\
no margin clipping & 19.57 & 98.24 & 37.00 & 1.48 & -7.02 & -0.35 \\
no target mse loss & 39.65 & 94.54 & 46.00 & -2.91 & -8.31 & -3.17 \\
strong target mse loss & 69.68 & 55.86 & 100.00 & 1.18 & -1.50 & 0.07 \\
\end{longtable}

We find that setting the clip\_epsilon\_logits parameter (the margin clip point that switches off preference loss for winning logits) to 16 -- effectively disabled -- results in model collapse. Logits diverge much further from reference, and output degrades to single-word repetitions. With this parameter set to 2 (the default), the model reaches the 95\% preference accuracy stopping point with writing quality preserved.

With lambda\_mse\_target reduced to 0, disabling the reference tether for target logits, we observe faster training and logits diverging farther from reference. Writing quality degrades 71\%  from the baseline per our rubric, illustrating the protective effect of this loss component. When lambda\_mse\_target is set to 0.4, logits diverged much less from reference, but the model was only able to achieve 74\% preference accuracy by training completion. At the default value of 0.05, the model reached the 95\% preference accuracy target without any substantial output degradation.

\clearpage
\section{DPO $\beta$ Hyperparameter Ablation}

\begin{figure}[h]
\begin{center}
\includegraphics[width=0.43\linewidth]{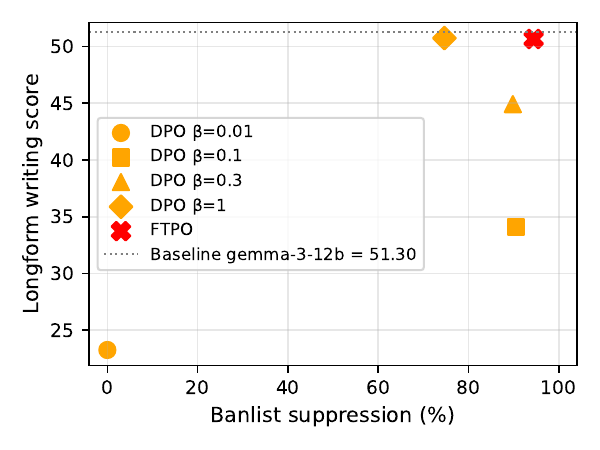}
\end{center}
\caption{We examine the impact of DPO's $\beta$ hyperparameter, training gemma-3-12b on our final-token preference set with several values of $\beta$: 0.01, 0.1, 0.3 and 1.0. This training set suppresses a banlist of 1,000 items. With DPO, we observe an expected tradeoff in learnability vs degradation \citep{wu2024betaDPO}. DPO manages a $<1\%$ reduction in output quality at $\beta=1.0$, but at the expense of significantly impaired banlist suppression (74.7\%). At lower values of $\beta$, output quality is markedly reduced for the DPO-trained models. In comparison, the FTPO model trained on the same dataset achieves the highest suppression rate of 94.4\% suppression, with neglibible ($<1\%$) degradation in longform writing score. }
\label{fig:dpo-beta-ablation}
\end{figure}

\section{Suppression Performance vs Writing Quality for EQ-Bench Dataset}

\begin{figure}[h]
\begin{center}
\includegraphics[width=0.46\linewidth]{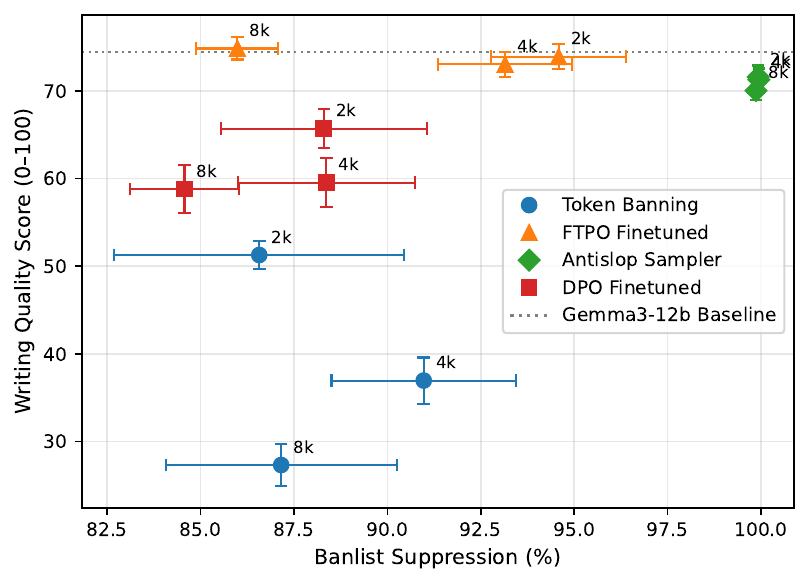}
\end{center}
\caption{We replicate \ref{sec:suppression-vs-quality} with an out-of-distribution writing prompts dataset. While a smaller dataset size of 96 prompts (and correspondingly larger error bars), we observe a similar pattern of banlist suppression rates and impact on writing quality for each method. }
\label{fig:quality_vs_banlist_eqbench}
\end{figure}

\clearpage
\section{Most Common Over-Represented Words and Trigrams Across Models}
\begin{table}[ht]
\centering
\begin{tabular}{ll}
\hline
\textbf{pattern} & \textbf{percent models} \\
\hline
flickered & 98.5 \\
flicker & 94.0 \\
flickering & 92.5 \\
leaned & 82.1 \\
muttered & 82.1 \\
gaze & 80.6 \\
grinned & 80.6 \\
containment & 77.6 \\
gestured & 77.6 \\
addendum & 74.6 \\
murmured & 73.1 \\
nodded & 73.1 \\
glint & 68.7 \\
hesitated & 68.7 \\
whispered & 68.7 \\
blinked & 64.2 \\
hummed & 64.2 \\
faintly & 62.7 \\
leans & 62.7 \\
unreadable & 62.7 \\
\hline
\end{tabular}
\caption{Top overlapping words across 67 AI models. Each entry shows the \% of models in which the token appears among their top 120 most over-represented words (relative to a human baseline).}
\label{tab:ai_overlap_words}
\end{table}

\begin{table}[ht]
\centering
\begin{tabular}{ll}
\hline
\textbf{pattern} & \textbf{percent models} \\
\hline
voice barely whisper & 68.7 \\
said voice low & 61.2 \\
air thick scent & 49.3 \\
took deep breath & 44.8 \\
smile playing lips & 43.3 \\
something else something & 37.3 \\
said voice barely & 35.8 \\
voice barely audible & 35.8 \\
take deep breath & 32.8 \\
could shake feeling & 31.3 \\
eyes never leaving & 29.9 \\
casting long shadows & 28.4 \\
says voice low & 26.9 \\
something else entirely & 26.9 \\
heart pounding chest & 25.4 \\
one last time & 23.9 \\
spreading across face & 22.4 \\
air thick smell & 19.4 \\
could help feel & 19.4 \\
long shadows across & 19.4 \\
\hline
\end{tabular}
\caption{Top overlapping trigrams across 67 AI models. Each entry shows the \% of models in which the phrase appears among their top 40 most over-represented trigrams (relative to a human baseline).}
\label{tab:ai_overlap_trigrams}
\end{table}

\clearpage
\section{Writing Quality Rubric Prompt}
\begin{center}
\begin{minipage}{\linewidth}
\centering
{\small
\begin{cfgblock}
You are an expert in assessing creative writing. Your task is to score the test model's response below, by several metrics, on a 0-20 scale.

[PROMPT START]
{writing_prompt}
[PROMPT END]

[TEST MODEL RESPONSE]
{test_model_response}
[TEST MODEL RESPONSE END]

[Task]

You are an expert in assessing creative writing. Your task is to score the model's response below, by several metrics, on a 0-20 scale.

Scoring notes:
- In the output, write the metric names exactly as below so they can be parsed.
- Use the designated output format exactly.
- All criteria are "higher is better"
- You are a critic, and your job is to be critical, especially of any failings or amateurish elements.
- Output format is:

[Scores]

Metric 1 name: [Score 0-20]

Metric 2 name: ...

---

Now, rate the supplied model output on the following criteria:

Spelling/grammar
Formatting issues & artifacts
Coherence
Consistency of tense, pronouns, perspective
Repetition issues
Overall quality of the piece
\end{cfgblock}
}
\captionof{figure}{Writing quality rubric prompt: This prompt was used to assess the overall quality of creative writing outputs in our experiments, with a particular focus on the common modes of degradation.}
\label{fig:writing-quality-rubric}
\end{minipage}
\end{center}

\clearpage
\section{Impact on Metrics by Banlist Size}

\begin{figure}[h]
\begin{center}
\includegraphics[width=0.85\linewidth]{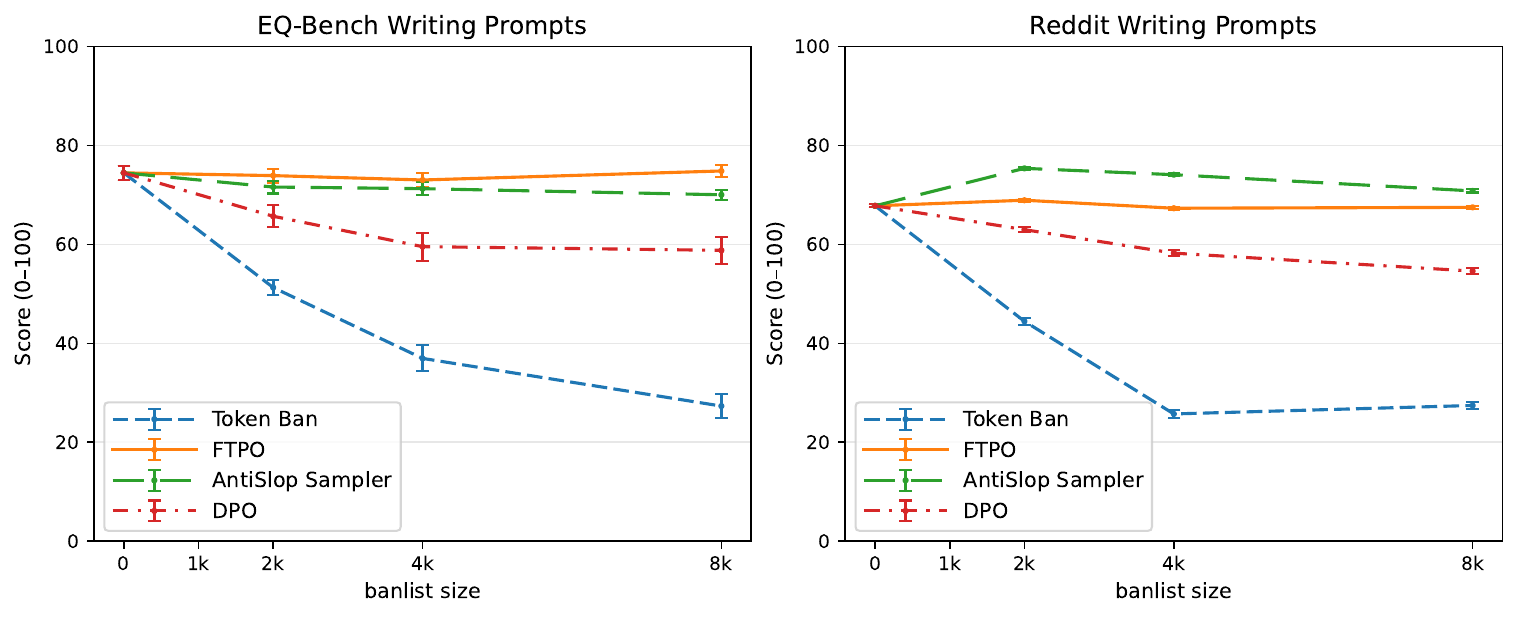}
\end{center}
\caption{Impact on writing quality per our LLM-judged rubric at several banlist sizes, for each suppression method (Token banning, FTPO, Antislop Sampler and DPO). }
\label{fig:four-methods-writing-quality}
\end{figure}

\begin{figure}[h]
\begin{center}
\includegraphics[width=0.85\linewidth]{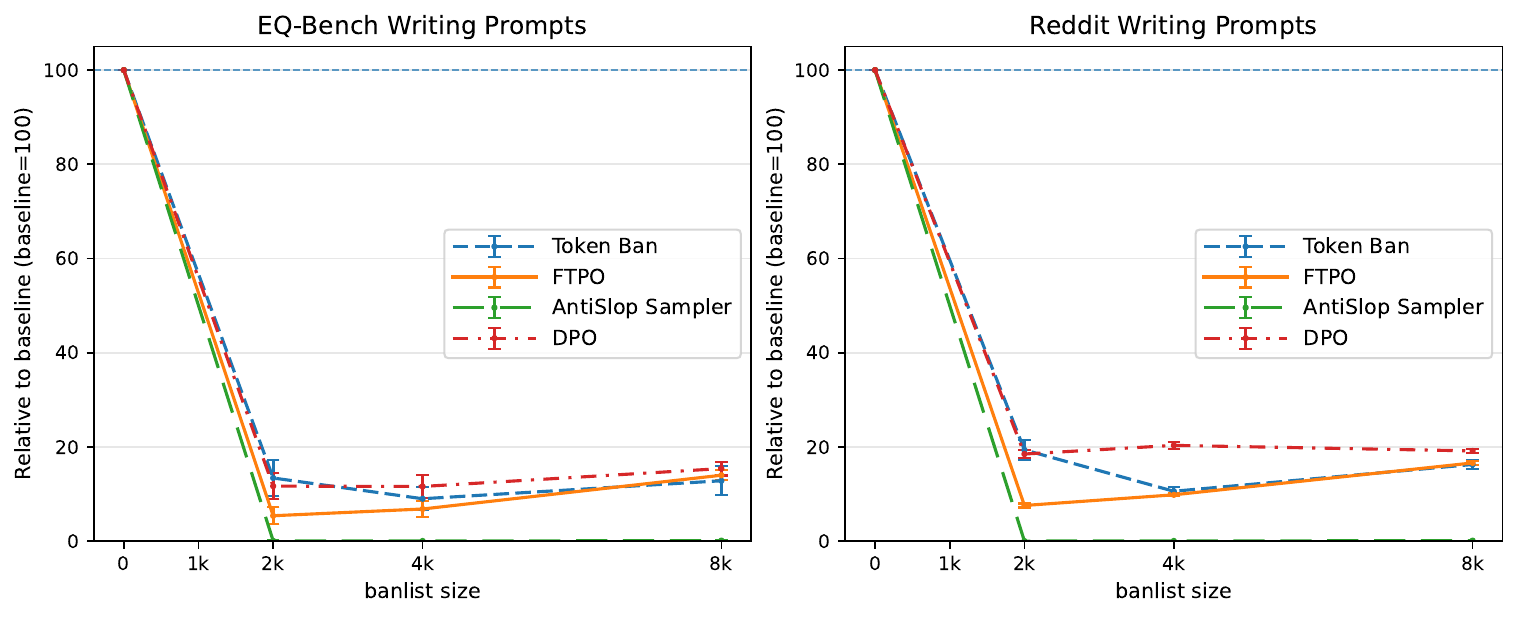}
\end{center}
\caption{Impact on banlist suppression rates at several banlist sizes, for each suppression method (Token banning, FTPO, Antislop Sampler and DPO). }
\label{fig:four-methods-banlist-suppression}
\end{figure}

\begin{figure}[h]
\begin{center}
\includegraphics[width=0.85\linewidth]{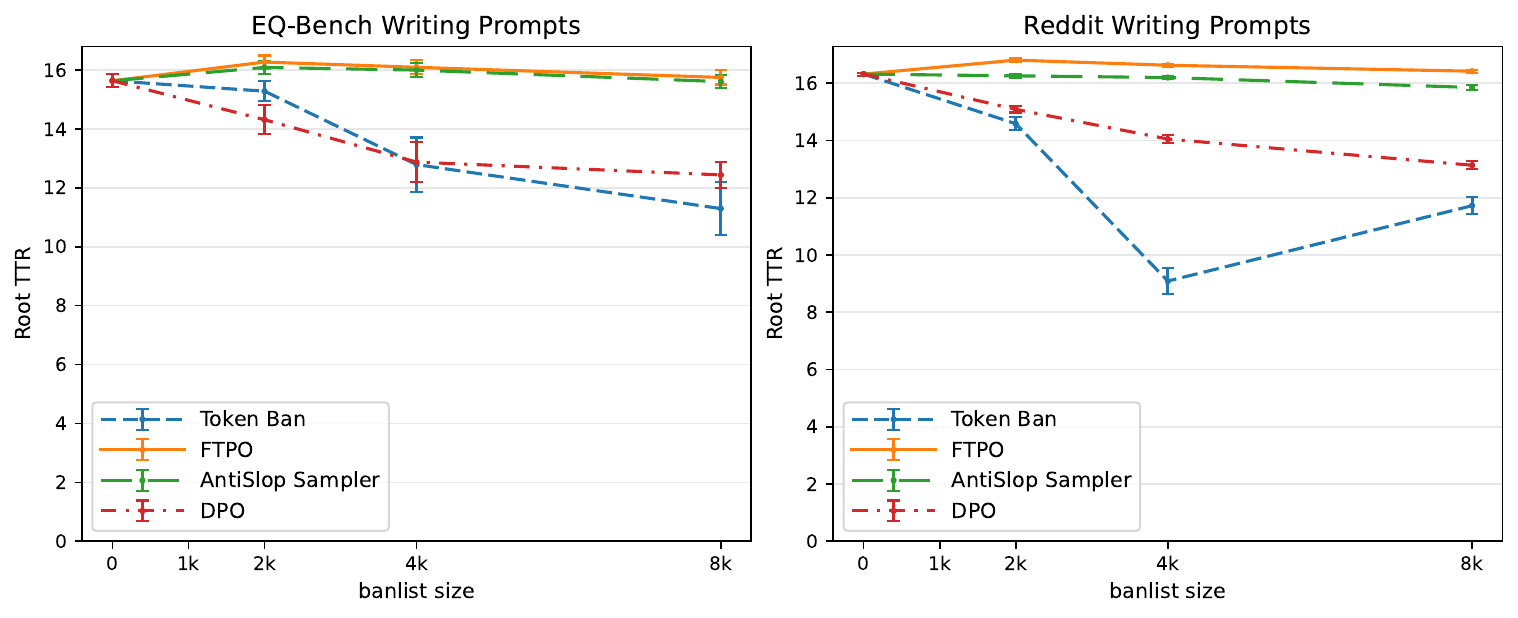}
\end{center}
\caption{Impact on lexical diversity at several banlist sizes, for each suppression method (Token banning, FTPO, Antislop Sampler and DPO). }
\label{fig:four-methods-banlist-suppression}
\end{figure}

\clearpage
\section{FTPO Loss Function Definition}

\textbf{Preference Loss Component:}

For each chosen token index $c$ against a rejected token index $r$, define the logit gap
\[
\Delta = y[c] - y[r].
\]

The margin requirement is $m$. A smooth penalty is applied if the gap is smaller than $m$:
\[
\ell^{\text{pref}} = \log\!\bigl(1 + e^{(m-\Delta)}\bigr),
\]
A taper weight
\[
w = \operatorname{clamp}\!\Big(\tfrac{m-\Delta}{m},\,0,\,1\Big)
\]
shrinks the contribution as $\Delta$ approaches the margin. The preference loss is the weighted mean over chosen tokens:
\[
\mathcal{L}_{\text{pref}} \;=\; \frac{\sum w\,\ell^{\text{pref}}}{\sum w}.
\]

\begin{figure}[h]
\begin{center}
\includegraphics[width=0.8\linewidth]{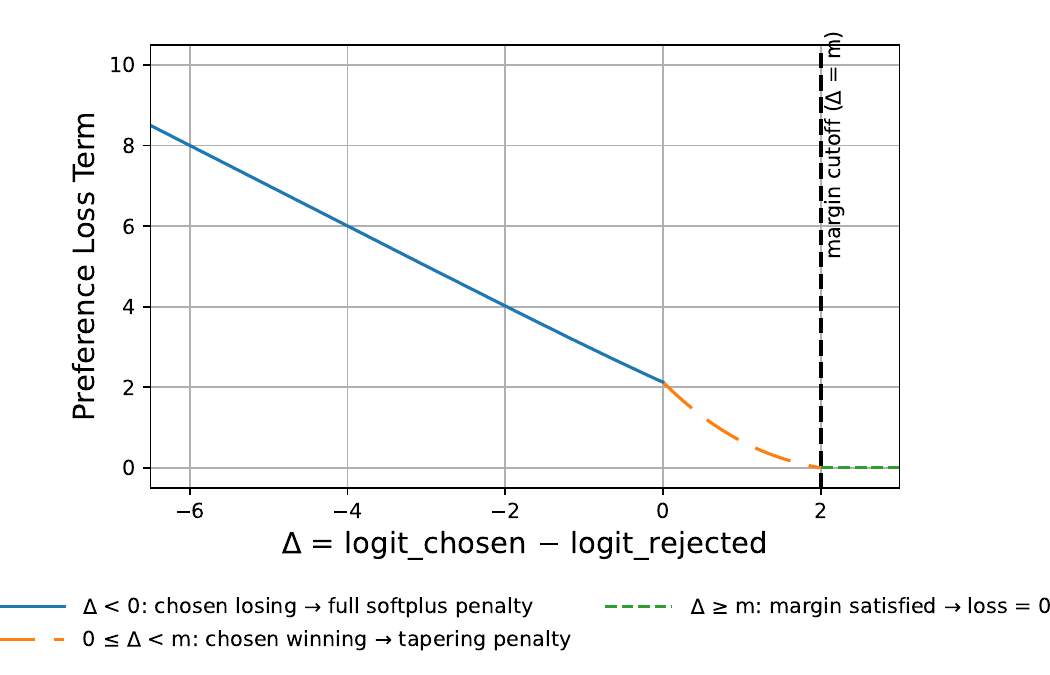}
\end{center}
\caption{Preference loss component as a function of the logit gap $\Delta$. When $\Delta < 0$ (chosen losing), the penalty is large. As $\Delta$ increases toward the margin $m$, the penalty smoothly tapers. Once $\Delta \ge m$, the weight goes to zero and the preference loss no longer contributes.}
\label{fig:pref-loss-component}
\end{figure}

\textbf{MSE tether terms:}

Let deviations be $d_j = y[j] - y^{\mathrm{ref}}[j]$. Define:
\begin{itemize}
    \item \textbf{Target set} $T = \{c\}\cup\{r\}$ (chosen and rejected indices).
    \item \textbf{Non-target set} $N = \{1,\dots,V\}\setminus T$.
\end{itemize}

\textbf{Non-target MSE loss term:}
\[
\mathcal{L}_{\text{nontarget}} = \frac{\sum_{j\in N} d_j^{\,2}}{|N|}.
\]

\textbf{Target MSE loss term with zero-penalty window}
\[
e_j = \max\!\bigl(|d_j| - \tau_{\text{target}},\,0\bigr), \qquad
\mathcal{L}_{\text{target}} = \frac{\sum_{j\in T} e_j^{\,2}}{|T|}.
\]

Here $\tau_{\text{target}}$ is a zero-penalty window: if the chosen or rejected logits are within $\pm \tau_{\text{target}}$ of the reference, no penalty is applied.

\begin{figure}[h]
\begin{center}
\includegraphics[width=0.8\linewidth]{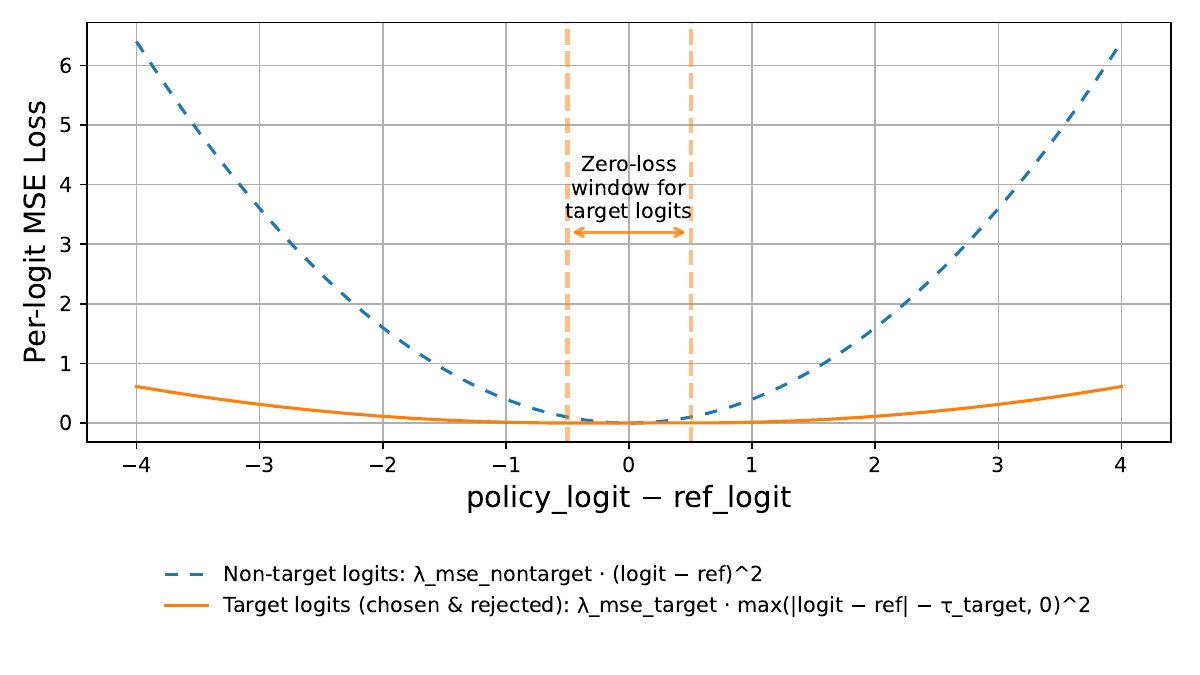}
\end{center}
\caption{MSE loss components as functions of logit deviation from the reference. The non-target term (blue) penalizes any deviation quadratically. The target term (orange) allows a dead zone around zero, where no penalty applies, then grows quadratically once the deviation exceeds the zero-penalty window.}
\label{fig:mse-loss-component}
\end{figure}

\textbf{Total objective:}

With weighting coefficients $\lambda_{\text{nontarget}}$ and $\lambda_{\text{target}}$, the total FTPO loss is
\[
\mathcal{L}
=
\mathcal{L}_{\text{pref}}
+
\lambda_{\text{nontarget}}\,\mathcal{L}_{\text{nontarget}}
+
\lambda_{\text{target}}\,\mathcal{L}_{\text{target}}.
\]

This formulation allows the model to learn a clear preference signal while preventing uncontrolled drift of the logit distribution.

\clearpage
\section{Slop Profile Clustering Between Models}
\label{app:slop-clustering}

\begin{wrapfigure}{r}{0.5\textwidth}
\vspace{-\baselineskip}
\centering
\includegraphics[width=\linewidth]{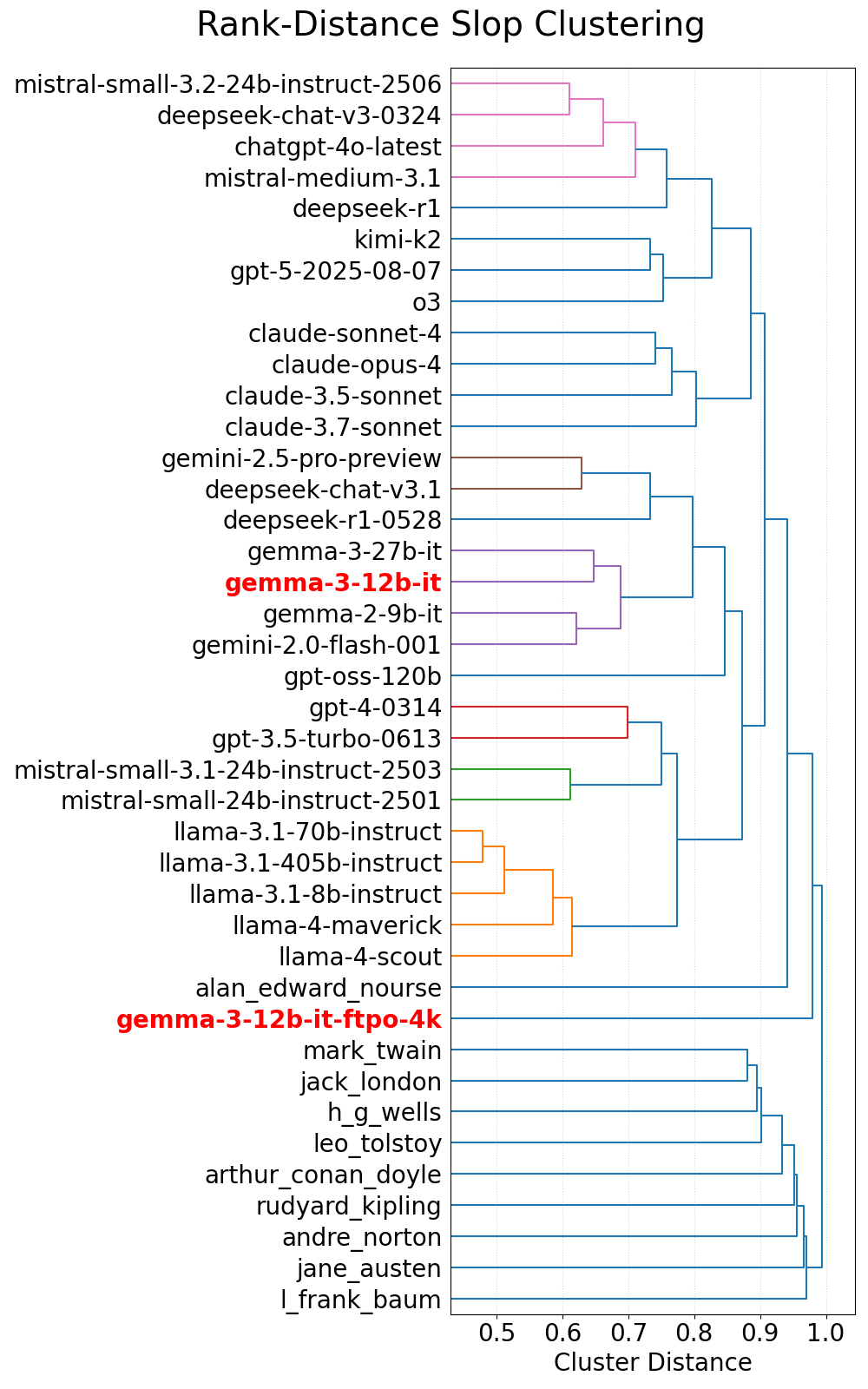}
\caption{Top 200 over-represented words and bigrams/trigrams were extracted for each model relative to a human baseline, for a set of creative writing outputs. For included human authors, a selection of their works were used. A dendrogram was generated with cluster distance as the \textbf{average ranking distance} of the top over-represented words \& n-grams list between models. Our FTPO antislop finetune of gemma-3-12b is highlighted, clustering closer to human authors than any other tested model.}
\label{fig:quant-slop-dendro}
\end{wrapfigure}

Colloquially, slop may refer to over-used words, phrases, themes or writing styles. Here we focus on over-used words and n-grams as they are relatively straightforward to extract. For a given model, we generate outputs from a creative writing prompts dataset \citep{paech2023eqbench} and a writing prompts dataset sourced from Reddit \citep{nitral_reddit_sfw_wp_sharegpt}. We then compute a list of the most over-represented words and bigrams/trigrams relative to a human baseline. The human baseline we use for individual words is the Python library \textbf{wordfreq} \citep{robyn_speer_2018_1443582}. For bigrams/trigrams, we compute a human baseline from a mix of sources including a large Reddit creative writing dataset, and a selection of public domain works from the Gutenberg Library \citep{project_gutenberg}. For n-gram extraction, we remove stop-words.

A "slop fingerprint" is collated from the top 120 most over-represented words and the top 40 most over-represented bigrams and trigrams. To avoid over-indexing on high-frequency words \& phrases in single texts (e.g. a character name), we require the pattern to occur from at least 3 writing prompts independently. To examine the relationship of this fingerprint between models, we perform hierarchical clustering on these top-200 lists per the average rank-distance between each model pair (Figure~\ref{fig:quant-slop-dendro}).

It's important to distinguish between counting the frequency of words and n-grams in a text, and calculating their frequency \textit{relative to a human baseline,} as we are doing here. The former simply surfaces patterns that are common in writing; the latter surfaces repetitive writing tendencies of a model that begin to stand out across multiple generations, leading to the perception of "slop". In some models this repetition is extreme: \textit{mistral-small-3.1-24b-instruct-2503} produced 102 \textit{"eyes never leaving"} trigrams and 62 \textit{"voice barely whisper"} trigrams across just 96 writing prompts.

We find a high correlation in words and n-grams found on the top most over-represented lists across the models tested, with \textit{"flickered"} appearing on 98.5\% of lists, and the trigram \textit{"voice barely whisper"} appearing on 68.7\% of lists. See Table~\ref{tab:ai_overlap_words} for the most commonly co-occurring word patterns across slop fingerprints, and Table~\ref{tab:ai_overlap_trigrams} for trigram patterns.

We utilise this method for identifying over-represented usages to compile a target list for slop reduction with the Antislop Sampler and FTPO fine-tuning. It should be noted that this method of identifying slop is domain-specific; the over-used patterns in creative writing will differ from professional writing, for instance. Here, we focus on creative writing, however the method can be applied to other domains by choosing a different set of prompts from which to derive the slop list.

\section{Regex blocklist used for ``not $x$, but $y$''}
\label{app:notxbuty-regex}
\begin{cfgblock}
regex_patterns: [
  
  "\\b(?:\\w+n(?:['’]t)|not\\s+(?:just|only|merely|because))\\s+(?:(?![.;:?!…]).){1,100}?[.;:?!…]\\s*(?:it|they|you)(?:['’](?:s|re|m))?\\b(?!\\s+(?:was|were|is|are|wasn['’]t|weren['’]t|isn['’]t|aren['’]t|ain['’]t)\\b)(?:\\s*[*…]?\\s*)?(?!when\\b|then\\b|but\\b|and\\b|yet\\b)(?!right\\b)(?!normal\\b)(?!true\\b)(?!sure\\b)(?!only\\b)(?!still\\b)(?!rarely\\b)(?!already\\b)(?!wrong\\b)(?!want\\b)(?!just\\b)(?!couldn\\b)(?!could\\b)(?!saw\\b)(?!started\\b)(?!remember\\b)(?!struggled\\b)(?!watched\\b)(?!goal\\b)(?!took\\b)(?!kept\\b)(?!reminded\\b)(?!time\\b)(?!have\\b)(?!acted\\b)(?!smiled\\b)(?!think\\b)(?!give\\b)(?!grab\\b)(?!gave\\b)(?!turn\\b)(?!justify\\b)(?!\\w+ly\\b)(?=[a-z]{4,}\\b)[a-z]+\\w*",

  "\\b(?:\\w+n(?:['’]t)|not)\\s+(?:just|only|merely)?\\s*(?:(?![-–—]|[.?!…]).){1,80}?[-–—]{1,2}\\s*\\w+(?:['’]\\w+)?\\s+",

  "\\b(?:wasn['’]t|weren['’]t|isn['’]t|aren['’]t|ain['’]t|not)\\s+(?!\\b(?:minute|minutes|hour|hours|day|days|year|years|second|seconds)\\b)(?!with\\b)(?!even\\b)(?:(?![.;:?!…]).){2,120}?[.;:?!…]\\s*(?:it|they|you|that)(?:\\s+(?:was|were|is|are)\\b(?:\\s+[*_~]?\\w+[*_~]?)?|(?:['’](?:s|re|m))\\b(?:\\s+[*_~]?\\w+[*_~]?)?)",

  "\\bno\\s+longer\\s+(?:just|only|merely)?\\s+[^.;:?!…]{1,120}[.;:?!…]\\s*(?:it|they|you)\\s+(?:is|are|was|were)\\b(?:\\s+[*_~]?\\w+[*_~]?)?",

  "\\b(?:wasn['’]t|weren['’]t|isn['’]t|aren['’]t|ain['’]t|not)\\s+(?:just|only|merely)?\\s*(?:(?!\\bbut\\b|[.?!…]).){1,80}?[,;:\\-–—]\\s*but\\s+(?!I\\b)(?:also\\s+)?"
]
\end{cfgblock}

\clearpage
\section{Auto-antislop Configuration File for gemma-3-12b-it 2k Banlist Size}
\label{app:gemma-config}

\begin{cfgblock}
    

####################################################################
# MAIN AUTO-ANTISLOP CONFIGURATION
####################################################################

####################################################################
# RUN SETUP
####################################################################
experiment_base_dir: "results/auto_antislop_runs" # Base for timestamped run directories
human_profile_path: "data/human_writing_profile.json"
log_level: "INFO"
# Iteration 0: Generates the baseline dataset & computes slop strings/ngrams to ban
# Iteration 1: Generates a dataset using antislop, banning those strings & ngrams. Recomputes the slop strings/ngrams at the end & adds any new slop to the banlists
# Iteration 2+: Extra iterations catch slop that emerges after the initial set is banned
num_iterations: 2 # Minimum 2 iterations (this is enough to catch most slop)
model_id: "google/gemma-3-12b-it" # Global model id for the pipeline. Can be overridden on individual steps.

####################################################################
# VLLM SERVER MANAGEMENT (Conditional: if --manage-vllm is True)
####################################################################
manage_vllm: true
vllm_model_id: null # Model served by vLLM (if unset, will use model_id)
vllm_port: 8000
vllm_hf_token: null # Optional: Your Hugging Face token if model is gated
vllm_cuda_visible_devices: "0"  # set to e.g. "0,1,2,3" for multiple gpus
vllm_gpu_memory_utilization: 0.85 # leave some room for the refusal classifier if you are using it (about 3gb)
vllm_max_model_len: 4500
vllm_dtype: "bfloat16"
# Additional raw CLI arguments for vLLM server, e.g., ["--tensor-parallel-size", "4"] for multiple gpus
vllm_extra_args: [] # each param as a separate string, e.g. ["--quantization", "bitsandbytes"]
vllm_env:                  # env vars for the vLLM process
  # VLLM_USE_V1: "1"  # may be needed for amd gpus


####################################################################
# GENERATION PARAMETERS (using antislop-vllm)
####################################################################
generation_step_enabled: true

# --- API & Model Configuration ---
# If you set manage_vllm=true, leave the base url unset
#generation_api_base_url: "http://localhost:8000/v1"
#generation_api_base_url: "https://apjmbtwbrb8t61-8888.proxy.runpod.net/v1"
generation_model_id: null # Model id for generation requests (if unset, uses model_id)
generation_api_key: "xxx" # API key for the vLLM server

# --- Core Generation Settings ---
generation_max_new_tokens: 1000
generation_threads: 50 # Number of parallel threads for API queries in antislop-vllm. Note: vllm can become very inefficient if you go over some concurrency threshold (depending on vram)
generation_max_prompts: 2000 # Number of samples to generate from the prompts in the dataset

# --- Dataset & Chat Template ---
generation_hf_dataset_name: 'Nitral-AI/Reddit-SFW-Writing_Prompts_ShareGPT'
generation_hf_dataset_split: 'train'
# A huggingface model id or local dir containing the tokeniser you want to use to apply chat templates.
# This is important if you are generating a ftpo dataset for later training.
generation_chat_template_model_id: null # (if unset, uses model_id)

# --- Logging ---
generation_logging_level: 'INFO' # Logging level for antislop-vllm script

# --- Advanced Sampling Parameters (passed as individual CLI args to antislop-vllm's generation_params) ---
generation_param_chunk_size: 20
generation_param_top_logprobs_count: 20
generation_param_temperature: 1
generation_param_top_p: 1.0
generation_param_top_k: 50
generation_param_min_p: 0.01
generation_param_timeout: 480
generation_param_stop_sequences: [] # e.g., ["\n\n", "---"]

# --- Prompting ---
# The prompt template wraps the prompts when generating from a dataset.
# To use the original prompt exactly, set the template to "{prompt}"
generation_prompt_template: "Writing prompt: {prompt}\n\nWrite 1000 words to this prompt. Your response:\n"
generation_system_prompt: "You are an uncensored writer."        # optional; left empty → no system prompt

# --- Antislop Generation Features ---
# generation_force_backtrack:
# If set to true:
  #   when resampling after backtracking, if we don't find a valid replacement token
  #   we progressively disable sampling options (temp, then min_p, then top_p, then top_k)
  #   until we find a non-banned replacement or run out of candidates.
  #   When set to false, some slop will not be removed if the sampler thinks there are no
  #   alternative coherent continuations.
generation_force_backtrack: false

# --- N-gram Validator Settings (for antislop-vllm) ---
# N-gram banlist file is managed by auto-antislop's iterative process.
generation_ngram_remove_stopwords: true
generation_ngram_language: "english"

# --- Refusal Detection ---
# Detects refusals & doesn't include them in the training dataset. Uses about 3GB extra VRAM.
generation_refusal_detection: true

####################################################################
# N-GRAM ANALYSIS & BANNING (within auto-antislop)
####################################################################
enable_ngram_ban: true
min_word_len_for_analysis: 3 # Filters out words under this length in n-gram analysis

# --- N-gram Identification Thresholds ---
top_k_bigrams: 5000
top_k_trigrams: 5000

# --- N-gram Banning Quotas (per iteration) ---
# Bigrams
dict_bigrams_initial: 300     # How many of the top over-represented dictionary bigrams to
                              # ban in the first antislop iteration.
                              # "Dictionary" means the bigrams were also found in the human
                              # writing corpus.
dict_bigrams_subsequent: 0   # How many to ban in each subsequent iteration
nodict_bigrams_initial: 200   # "Nodict" here means the n-grams were not found at all in the
                              # human corpus.
nodict_bigrams_subsequent: 0
# Trigrams
dict_trigrams_initial: 300
dict_trigrams_subsequent: 0
nodict_trigrams_initial: 200
nodict_trigrams_subsequent: 0

# --- User-Defined N-gram Bans ---
# User-supplied extra n-grams to always ban (processed by auto-antislop)
extra_ngrams_to_ban: [
  # "voice barely whisper",
]

####################################################################
# OVER-REPRESENTED WORD ANALYSIS & BANNING
####################################################################
compute_overrep_words: true
top_k_words_for_overrep_analysis: 200000

# --- Quotas for Adding Over-represented Words to Slop Phrase banlist ---
dict_overrep_initial: 920       # How many of the top over-represented dictionary words to
                                # ban in the first antislop iteration.
                                # "Dictionary" means the words were also found in the human
                                # writing corpus.
dict_overrep_subsequent: 0    # How many to ban in each subsequent iteration
nodict_overrep_initial: 80      # "Nodict" here means the n-grams were not found at all in the
                                # human corpus.
nodict_overrep_subsequent: 0

####################################################################
# SLOP PHRASE BANNING
####################################################################

# Slop phrases are over-represented whole phrases extracted from the generated texts.
enable_slop_phrase_ban: true
min_phrase_freq_to_keep: 2 # Min frequency for a new phrase from slop-forensics to be considered
top_n_initial_slop_ban: 0 # New slop phrases from slop-forensics to ban in iter 0
top_n_subsequent_slop_ban: 0 # New slop phrases from slop-forensics to ban in later iters

# --- User-Defined Slop Phrase Bans ---
# User supplied list of strings to always ban
# - case insensitive
# To trigger a ban, the sequence must not have a word-like character
#    (not punctuation or whitespace) directly on either side. That is to say, we
#    are not banning disallowed sequences that occur as substrings in longer
#    words. The exception is if the banned string is already bookended by
#    a non-word character.
#
#    Examples:
#    banned string "cat"
#      - won't trigger a ban for "cation"
#     - will trigger a ban on "cat[morecat]"
#   banned string "cat["
#     - *will* trigger a ban on "cat[morecat]", because the banned string
#        ends with a non-word character.
extra_slop_phrases_to_ban: [
  # "…", "...", "rain", "tapestry", "static", "regret", "rust"
]

# --- Whitelisted Strings ---
# These will be excluded from the list of slop strings that the pipeline finds.
# Note: special tokens in the tokenizer and parts of the chat template are
#       automatically whitelisted.
whitelist_strings: [
  # "think", "thinking"
]

####################################################################
# REGEX BANNING
####################################################################
# User-supplied regex patterns to ban
# Note: unoptimised regex patterns can slow down antislop generation, as they will be called often on large texts.
extra_regex_patterns: [
  # These ones ban "it's not x, it's y" type patterns:
    
  #"\\b(?:\\w+n(?:['’]t)|not\\s+(?:just|only|merely|because))\\s+(?:(?![.;:?!…]).){1,100}?[.;:?!…]\\s*(?:it|they|you)(?:['’](?:s|re|m))?\\b(?!\\s+(?:was|were|is|are|wasn['’]t|weren['’]t|isn['’]t|aren['’]t|ain['’]t)\\b)(?:\\s*[*…]?\\s*)?(?!when\\b|then\\b|but\\b|and\\b|yet\\b)(?!right\\b)(?!normal\\b)(?!true\\b)(?!sure\\b)(?!only\\b)(?!still\\b)(?!rarely\\b)(?!already\\b)(?!wrong\\b)(?!want\\b)(?!just\\b)(?!couldn\\b)(?!could\\b)(?!saw\\b)(?!started\\b)(?!remember\\b)(?!struggled\\b)(?!watched\\b)(?!goal\\b)(?!took\\b)(?!kept\\b)(?!reminded\\b)(?!time\\b)(?!have\\b)(?!acted\\b)(?!smiled\\b)(?!think\\b)(?!give\\b)(?!grab\\b)(?!gave\\b)(?!turn\\b)(?!justify\\b)(?!\\w+ly\\b)(?=[a-z]{4,}\\b)[a-z]+\\w*",

  #"\\b(?:\\w+n(?:['’]t)|not)\\s+(?:just|only|merely)?\\s*(?:(?![-–—]|[.?!…]).){1,80}?[-–—]{1,2}\\s*\\w+(?:['’]\\w+)?\\s+",

  #"\\b(?:wasn['’]t|weren['’]t|isn['’]t|aren['’]t|ain['’]t|not)\\s+(?!\\b(?:minute|minutes|hour|hours|day|days|year|years|second|seconds)\\b)(?!with\\b)(?!even\\b)(?:(?![.;:?!…]).){2,120}?[.;:?!…]\\s*(?:it|they|you|that)(?:\\s+(?:was|were|is|are)\\b(?:\\s+[*_~]?\\w+[*_~]?)?|(?:['’](?:s|re|m))\\b(?:\\s+[*_~]?\\w+[*_~]?)?)",

  #"\\bno\\s+longer\\s+(?:just|only|merely)?\\s+[^.;:?!…]{1,120}[.;:?!…]\\s*(?:it|they|you)\\s+(?:is|are|was|were)\\b(?:\\s+[*_~]?\\w+[*_~]?)?",

  #"\\b(?:wasn['’]t|weren['’]t|isn['’]t|aren['’]t|ain['’]t|not)\\s+(?:just|only|merely)?\\s*(?:(?!\\bbut\\b|[.?!…]).){1,80}?[,;:\\-–—]\\s*but\\s+(?!I\\b)(?:also\\s+)?"

]

####################################################################
# FINETUNING
####################################################################
finetune_enabled: true

# --- General Finetuning Setup ---
finetune_use_unsloth: false
finetune_mode: "ftpo" # ftpo | dpo-final-token (final token preference optimisation)
finetune_ftpo_dataset: ""   # you can specify an existing ftpo dataset, or leave unset to let the
                            # pipeline use the one produced in the generation step
finetune_base_model_id: null # Base model for DPO (if unset, uses model_id)
finetune_max_seq_length: 2500 # this may truncate some outputs
finetune_load_in_4bit: true # qlora

# --- Early Stopping ---
finetune_early_stopping_wins: 0.85  # Early stopping threshold for fraction of *chosen* completions that are selected over *rejected*.
                                    # More than 0.85 may be overtrained. Set to > 1.0 to disable early stopping.
finetune_early_stopping_loss: null  # Loss threshold for early stopping. Set to null to disable.

# --- LoRA Configuration ---
finetune_lora_r: 256 # the ftpo trainer works best with a high lora rank
finetune_lora_alpha: 256
finetune_lora_dropout: 0.05
finetune_weight_decay: 0.01
finetune_target_modules: ["up_proj", "down_proj", "lm_head"]

# --- Layer Freezing ---
finetune_freeze_early_layers: true
finetune_n_layers_unfrozen: 5

# --- Training Process ---
finetune_gradient_checkpointing: "unsloth"
finetune_chat_template: "" # e.g. "gemma-3" -- get the chat template from unsloth's helper if required, otherwise leave the string blank to use the tokeniser's chat template
finetune_batch_size: 3
finetune_gradient_accumulation_steps: 5
finetune_warmup_ratio: 0.1
finetune_num_epochs: 1

# --- Learning Rate ---
finetune_learning_rate: 0.000001
finetune_auto_learning_rate: true  # true: automatically determine learning rate based on dataset size, effective batch size & lora rank
finetune_auto_learning_rate_adjustment_scaling: 0.08 # scale the auto-lr by this factor

# --- DPO/FTPO Specific ---
finetune_beta: 0.1 # DPO beta

# --- Output & Saving ---
finetune_output_dir_suffix: "_ftpo_exp01" # Appended to experiment run dir
finetune_save_merged_16bit: true
finetune_save_gguf_q8_0: false

# --- Dataset Handling for Finetuning ---
finetune_max_train_examples: 12000 # adjust as needed
finetune_shuffle_seed: 42

# --- FTPO Sample Regularization ---
# 0 = off; 0.9 strongly downsamples overrepresented rule violations
# (this is useful because the raw generated dataset is typically very skewed)
ftpo_sample_rejected_regularisation_strength: 0.8
ftpo_sample_chosen_regularisation_strength: 0.2
ftpo_sample_min_chosen_tokens: 4 # filter out ftpo samples that have fewer than this number in the chosen tokens list


# FTPO-specific hyper-parameters
# Leave any of these out (or set to null) to fall back to FTPOTrainer defaults.

# Loss terms are computed separately for the target (chosen + rejected) tokens vs the remainder of the vocab.
# This is because we want to allow more freedom of movement for the target tokens.

# MSE loss term 1: light mse loss applied tokenwise on target tokens
ftpo_lambda_mse_target: 0.05   # Strength of MSE loss tether on the individual logits in the
                                        #   chosen+rejected set vs reference.
ftpo_tau_mse_target: 0.5       # Grace bandwidth (logits) before the above MSE loss kicks in.

# MSE loss term 2: stronger mse term applied to remaining (non-target) vocab
ftpo_lambda_mse: 0.4

ftpo_clip_epsilon_logits: 2     # For a chosen token: "after winning vs rejected token by this margin, preference loss turns off"

\end{cfgblock}

\end{document}